\documentclass{article}

 \usepackage[accepted]{aistats2022}

\usepackage[utf8]{inputenc} % allow utf-8 input
\usepackage[T1]{fontenc}    % use 8-bit T1 fonts
\usepackage{hyperref}       % hyperlinks
\usepackage{url}            % simple URL typesetting
\usepackage{booktabs}       % professional-quality tables
\usepackage{amsfonts}       % blackboard math symbols
\usepackage{nicefrac}       % compact symbols for 1/2, etc.
\usepackage{microtype}      % microtypography

\usepackage{microtype}
\usepackage{graphicx}
\usepackage{booktabs} % for professional tables

\usepackage{natbib}
\usepackage{amssymb, amsmath,amsthm}
\usepackage{amsfonts}
\usepackage{algorithm}
\usepackage{algorithmic}
\usepackage{paralist}
\usepackage{multirow}
\usepackage{times}
\usepackage{wrapfig}
\usepackage{url,enumerate}
\usepackage{color,xcolor}
\usepackage{makeidx}  % allows for indexgeneration
\usepackage{amsmath,amssymb}
\usepackage{mathtools}
\usepackage[small, compact]{titlesec}
\usepackage{xspace}
\usepackage{epstopdf}
\usepackage{mathrsfs}
\usepackage{times}
\usepackage{enumerate}
\usepackage{color}
\usepackage{graphicx,epsfig}
\usepackage{amsmath,amssymb,xspace}
\usepackage{url}
\usepackage{bm}
\usepackage{bbm}
\usepackage{upgreek}
\usepackage{cleveref}
\usepackage{multirow}
\usepackage{ulem}
\usepackage{cancel}
\usepackage{subcaption}
\usepackage{float}

\newcommand{\ours}{PI-DKL\xspace}

\begin{document}

\twocolumn[

\aistatstitle{Physics Informed Deep Kernel Learning}

\aistatsauthor{ Zheng Wang \And Wei Xing \And  Robert M. Kirby \And Shandian Zhe }

\aistatsaddress{  University of Utah \And  University of Utah  \And University of Utah \And Univeristy of Utah } ]

%\maketitle

% Math commands by Thomas Minka
\newcommand{\var}{{\rm var}}
\newcommand{\Tr}{^{\rm T}}
\newcommand{\vtrans}[2]{{#1}^{(#2)}}
\newcommand{\kron}{\otimes}
\newcommand{\schur}[2]{({#1} | {#2})}
\newcommand{\schurdet}[2]{\left| ({#1} | {#2}) \right|}
\newcommand{\had}{\circ}
\newcommand{\diag}{{\rm diag}}
\newcommand{\invdiag}{\diag^{-1}}
\newcommand{\rank}{{\rm rank}}
% careful: ``null'' is already a latex command
\newcommand{\nullsp}{{\rm null}}
\newcommand{\tr}{{\rm tr}}
\renewcommand{\vec}{{\rm vec}}
\newcommand{\vech}{{\rm vech}}
\renewcommand{\det}[1]{\left| #1 \right|}
\newcommand{\pdet}[1]{\left| #1 \right|_{+}}
\newcommand{\pinv}[1]{#1^{+}}
\newcommand{\erf}{{\rm erf}}
\newcommand{\hypergeom}[2]{{}_{#1}F_{#2}}

% boldface characters
\renewcommand{\a}{{\bf a}}
\renewcommand{\b}{{\bf b}}
\renewcommand{\c}{{\bf c}}
\renewcommand{\d}{{\rm d}}  % for derivatives
\newcommand{\e}{{\bf e}}
\newcommand{\f}{{\bf f}}
\newcommand{\g}{{\bf g}}
\newcommand{\h}{{\bf h}}
%\newcommand{\k}{{\bf k}}
% in Latex2e this must be renewcommand
\renewcommand{\k}{{\bf k}}
\newcommand{\m}{{\bf m}}
\newcommand{\mb}{{\bf m}}
\newcommand{\n}{{\bf n}}
\renewcommand{\o}{{\bf o}}
\newcommand{\p}{{\bf p}}
\newcommand{\q}{{\bf q}}
\renewcommand{\r}{{\bf r}}
\newcommand{\s}{{\bf s}}
\renewcommand{\t}{{\bf t}}
\renewcommand{\u}{{\bf u}}
\renewcommand{\v}{{\bf v}}
\newcommand{\w}{{\bf w}}
\newcommand{\x}{{\bf x}}
\newcommand{\hx}{{\hat{\x}}}
\newcommand{\hf}{{\hat{\f}}}

\newcommand{\y}{{\bf y}}
\newcommand{\z}{{\bf z}}
%s\newcommand{\l}{\boldsymbol{l}}
\newcommand{\A}{{\bf A}}
\newcommand{\B}{{\bf B}}
\newcommand{\C}{{\bf C}}
\newcommand{\D}{{\bf D}}
\newcommand{\E}{{\bf E}}
\newcommand{\F}{{\bf F}}
\newcommand{\G}{{\bf G}}
\renewcommand{\H}{{\bf H}}
\newcommand{\I}{{\bf I}}
\newcommand{\J}{{\bf J}}
\newcommand{\K}{{\bf K}}
\newcommand{\hK}{\widehat{\K}}
\renewcommand{\L}{{\bf L}}
\newcommand{\M}{{\bf M}}
\newcommand{\N}{\mathcal{N}}  % for normal density
\newcommand{\Acal}{\mathcal{A}}
\newcommand{\Ocal}{\mathcal{O}}
\newcommand{\Dcal}{\mathcal{D}}
\newcommand{\Ycal}{\mathcal{Y}}
\newcommand{\Zcal}{\mathcal{Z}}
\newcommand{\Fcal}{\mathcal{F}}
\newcommand{\Vcal}{\mathcal{V}}
\newcommand{\Lcal}{\mathcal{L}}
\newcommand{\Tcal}{\mathcal{T}}
\newcommand{\Gcal}{\mathcal{G}}
\newcommand{\Hcal}{\mathcal{H}}
\newcommand{\Scal}{\mathcal{S}}

\renewcommand{\O}{{\bf O}}
\renewcommand{\P}{{\bf P}}
\newcommand{\Q}{{\bf Q}}
\newcommand{\R}{{\bf R}}
\renewcommand{\S}{{\bf S}}
\newcommand{\T}{{\bf T}}
\newcommand{\U}{{\bf U}}
\newcommand{\V}{{\bf V}}
\newcommand{\W}{{\bf W}}
\newcommand{\X}{{\bf X}}
\newcommand{\hX}{{\hat{\X}}}
\newcommand{\Y}{{\bf Y}}
\newcommand{\Z}{{\bf Z}}
\newcommand{\Mcal}{{\mathcal{M}}}
\newcommand{\Wcal}{{\mathcal{W}}}
\newcommand{\Ucal}{{\mathcal{U}}}

% this is for latex 2.09
% unfortunately, the result is slanted - use Latex2e instead
%\newcommand{\bfLambda}{\mbox{\boldmath$\Lambda$}}
% this is for Latex2e
\newcommand{\bfLambda}{\boldsymbol{\Lambda}}

% Yuan Qi's boldsymbol
\newcommand{\bsigma}{\boldsymbol{\sigma}}
\newcommand{\balpha}{\boldsymbol{\alpha}}
\newcommand{\bpsi}{\boldsymbol{\psi}}
\newcommand{\bphi}{\boldsymbol{\phi}}
\newcommand{\boldeta}{\boldsymbol{\eta}}
\newcommand{\Beta}{\boldsymbol{\eta}}
\newcommand{\btau}{\boldsymbol{\tau}}
\newcommand{\bvarphi}{\boldsymbol{\varphi}}
\newcommand{\bzeta}{\boldsymbol{\zeta}}

\newcommand{\blambda}{\boldsymbol{\lambda}}
\newcommand{\bLambda}{\mathbf{\Lambda}}
\newcommand{\bOmega}{\mathbf{\Omega}}
\newcommand{\bomega}{\mathbf{\omega}}
\newcommand{\bPi}{\mathbf{\Pi}}

\newcommand{\btheta}{\boldsymbol{\theta}}
\newcommand{\bpi}{\boldsymbol{\pi}}
\newcommand{\bxi}{\boldsymbol{\xi}}
\newcommand{\bSigma}{\boldsymbol{\Sigma}}

\newcommand{\bgamma}{\boldsymbol{\gamma}}
\newcommand{\bGamma}{\mathbf{\Gamma}}

\newcommand{\bmu}{\boldsymbol{\mu}}
\newcommand{\1}{{\bf 1}}
\newcommand{\0}{{\bf 0}}

\newcommand{\bs}{\backslash}
\newcommand{\ben}{\begin{enumerate}}
\newcommand{\een}{\end{enumerate}}

 \newcommand{\notS}{{\backslash S}}
 \newcommand{\nots}{{\backslash s}}
 \newcommand{\noti}{{\backslash i}}
 \newcommand{\notj}{{\backslash j}}
 \newcommand{\nott}{\backslash t}
 \newcommand{\notone}{{\backslash 1}}
 \newcommand{\nottp}{\backslash t+1}

\newcommand{\notk}{{^{\backslash k}}}
\newcommand{\notij}{{^{\backslash i,j}}}
\newcommand{\notg}{{^{\backslash g}}}
\newcommand{\wnoti}{{_{\w}^{\backslash i}}}
\newcommand{\wnotg}{{_{\w}^{\backslash g}}}
\newcommand{\vnotij}{{_{\v}^{\backslash i,j}}}
\newcommand{\vnotg}{{_{\v}^{\backslash g}}}
\newcommand{\half}{\frac{1}{2}}
\newcommand{\msgb}{m_{t \leftarrow t+1}}
\newcommand{\msgf}{m_{t \rightarrow t+1}}
\newcommand{\msgfp}{m_{t-1 \rightarrow t}}

\newcommand{\proj}[1]{{\rm proj}\negmedspace\left[#1\right]}
\newcommand{\argmin}{\operatornamewithlimits{argmin}}
\newcommand{\argmax}{\operatornamewithlimits{argmax}}

\newcommand{\dif}{\mathrm{d}}
\newcommand{\abs}[1]{\lvert#1\rvert}
\newcommand{\norm}[1]{\lVert#1\rVert}

%miscellaneous symbols
%\newcommand{\ie}{{{\em i.e.,}}\xspace}
\newcommand{\ie}{{\textit{i.e.,}}\xspace}
\newcommand{\eg}{{\textit{e.g.,}}\xspace}
\newcommand{\etc}{{\textit{etc.}}\xspace}
\newcommand{\EE}{\mathbb{E}}
\newcommand{\dr}[1]{\nabla #1}
\newcommand{\VV}{\mathbb{V}}
\newcommand{\sbr}[1]{\left[#1\right]}
\newcommand{\rbr}[1]{\left(#1\right)}
\newcommand{\cmt}[1]{}
\newcommand{\tg}{\widetilde{g}}
\newcommand{\tZ}{\widetilde{\Z}}
\newcommand{\teps}{\widetilde{\epsilon}}

\newcommand{\bi}{{\bf i}}
\newcommand{\bj}{{\bf j}}
\newcommand{\bK}{{\bf K}}

\begin{abstract}
	Deep kernel learning is a promising combination of deep neural networks and nonparametric function learning. However, as a data driven approach,  the performance of deep kernel learning can still be restricted by scarce or insufficient data, especially in extrapolation tasks.  To address these limitations, we propose Physics Informed Deep Kernel Learning (\ours) that exploits physics  knowledge represented by differential equations with latent sources. Specifically, we use the posterior function sample of the  Gaussian process as the surrogate for the solution of the differential equation, and construct  a generative component to integrate the equation in a principled Bayesian hybrid framework. For efficient and effective inference, we marginalize out the latent variables in the joint probability and derive a collapsed model evidence lower bound (ELBO), based on which we develop a stochastic model estimation algorithm. Our ELBO can be viewed as a nice, interpretable posterior regularization objective. On synthetic datasets and real-world applications, we show the advantage of our approach in both prediction accuracy and uncertainty quantification. 
\end{abstract}
%deep kernel learning background --> issue in data sparsity and extrapolation --> physics modeling --> our method --> experiments
\section{Introduction}
%\vspace{-0.1in}
Deep kernel learning~\citep{wilson2016deep} uses deep neural networks to construct kernels for nonparametric function learning (typically via Gaussian processes~\citep{williams2006gaussian}) and unifies both the expressive power of neural networks and self-adaptation of nonparametric function estimators. Many applications have shown that deep kernel learning  substantially outperforms the conventional shallow kernel learning (\eg RBF). Compared to standard neural networks, deep kernel learning enjoys closed-form posterior (or predictive) distributions and hence is more convenient for uncertainty quantification and reasoning, which is important for decision making. 

Nonetheless, as a data driven approach, the performance of deep kernel learning can still be restricted by scarce data, especially when the training samples are insufficient to reflect the complexity of the system (that produced the data) or the test points are far away from the training set, \ie extrapolation. On the other hand,  physics knowledge, expressed as differential equations,  are used to build physical models for various science and engineering applications~\citep{lapidus2011numerical}. These models are meant to characterize the underlying mechanism (\ie physical processes) that drives the system (\eg how the heat diffuses across the spatial and temporal domains) and are much less restricted by data availability: they can make accurate predictions even without training data, \eg the landing of Curiosity on Mars and flight of Voyager 1. 

Therefore, we consider integrating physics knowledge into deep kernel learning to further improve its performance in prediction and uncertainty quantification,  especially for scarce data and extrapolation tasks. Our work is enlightened by the recent Physics Informed Neural Networks (PINNs) ~\citep{raissi2019physics}.  However, there are two substantial differences. First, PINNs require the form of the differential equation to be fully specified. We allow the equation to include unknown latent sources (functions), which is often the case in practice. Second, we integrate the differential equation in a principled Bayesian manner to pursue better calibrated posterior estimations. 

Specifically, we use the posterior sample of the Gaussian process (GP), which is a random function, as the surrogate for the solution of the differential equation. We then apply the differential operators in the equation to obtain the sample of the latent source (function), for which we assign  another GP prior. To ensure the sampling procedure is valid, we use the symmetric property of the Gaussian distribution to sample a set of virtual data $\{0\}$, which is computationally equivalent to placing the GP prior with a zero mean function over the latent source. The sampling procedure constitutes  a generative component and ties to the original deep kernel model in the Bayesian hybrid framework~\citep{lasserre2006principled}.  For efficient and high-quality inference, we marginalize out all the latent variables in the joint distribution to avoid approximating their complex posteriors, and use Jensen's inequality to obtain a collapsed model evidence lower bound (ELBO). We then  develop a stochastic model estimation algorithm. The ELBO can be further explained as a soft posterior regularization objective ~\citep{ganchev2010posterior}, regularized by physics.

For evaluation, we examined our physics informed deep kernel learning (\ours) in both simulation and real-world applications. 
On synthetic datasets based on two commonly used differential equations, \ours outperforms the standard deep kernel learning, shallow kernel learning, and latent force models (LFM) that combine the physics via kernel convolution, in both ground-truth function recovery and prediction uncertainty, especially in the case of extrapolation. We then examined \ours in four real-world applications, where two of them involve nonlinear differential equations with an unknown source function of both the time and spatial variables.  \ours consistently improves upon the competing approaches in prediction error and test log-likelihood. %We applied \ours with a nonlinear differential equation where LFM is infeasible. \ours significantly outperforms standard deep/shallow kernel learning methods.  

%\vspace{-0.1in}
\section{Background} \label{sec:background}
%\vspace{-0.1in}
%GP, LFM 
%\subsection{Standard Gaussian Process} \label{sec:standard-gaussian-process}
\noindent \textbf{Gaussian Process and Kernel Learning}. The Gaussian process (GP) is the most commonly used nonparametric function prior for kernel learning. Suppose we aim to learn a function $f: \mathbb{R}^d \rightarrow \mathbb{R}$ from a training set $\Dcal = (\X, \y)$, where  $\X=[\x_1, \cdots, \x_N]^\top$, $\y=[y_1, \cdots, y_N]^\top$, each $\x_n$ is a $d$ dimensional input vector and $y_n$ the observed output. To avoid both under-fitting and over-fitting, we do not want to assume any parametric form of $f$. Instead, we want the complexity of $f(\cdot)$ to automatically adapt to the data. To this end, we introduce a kernel function $k(\cdot, \cdot)$ that measures the similarity of the function outputs in terms of their inputs. The similarity only brings in a smoothness assumption about the target function. For example, the commonly used RBF kernel, $k_{\text{RBF}}(\x_i, \x_j) = \exp(-\frac{\|\x_i - \x_j\|^2}{\eta})$, implies the function is infinitely differentiable. We then use the kernel to construct a GP prior, $f\sim \mathcal{GP}\left(m(\cdot), k(\cdot, \cdot)\right)$ where $m(\cdot)$ is the mean function that is usually set to constant $0$. According to the GP definition, the finite projection of $f(\cdot)$ on the training inputs $\X$, namely $\f = [f(\x_1), \cdots, f(\x_N)]^\top$, follow a multivariate Gaussian distribution, $p(\f|\X) = \N(\f|\0, \K)$ where $\K$ is the kernel matrix on $\X$ and each  $[\K]_{i,j}=k(\x_i, \x_j)$.
Given the function values $\f$, the observed outputs $\y$ are sampled from a noisy model. For example, when $\y$ are continuous, we can use the isotropic Gaussian noise model, $p(\y|\f) = \N(\y|\f, \tau^{-1}\I)$ where $\tau$ is the inverse variance. We can integrate out $\f$ to obtain the marginal likelihood of $\y$, 
\begin{align}
	p(\y|\X) = \N(\y|\0, \K + \tau^{-1}\I). \label{eq:gp}
\end{align} 
To learn the model, we can maximize the likelihood to estimate the kernel parameters and the inverse variance $\tau$.  According to the GP prior, given a new input $\x^*$,  the posterior (or predictive) distribution of the output $f(\x^*)$ is a conditional Gaussian, 
\begin{align}
	p\big(f(\x^*)|\x^*, \X, \y\big) = \N\big(f(\x^*)|\mu(\x^*), v(\x^*)\big),
	\label{eq:predictive}
\end{align}
where $\mu(\x^*) = \k_*^\top(\K + \tau^{-1}\I)^{-1}\y$, $v(\x^*)=k(\x^*, \x^*) - \k_*^\top(\K + \tau^{-1}\I)^{-1}\k_*$ and  $\k_* = [k(\x^*, \x_1), \cdots, k(\x^*, \x_N)]^\top$. 

%multiple Q otput functions, each 
\noindent\textbf{Deep Kernel Learning}. While GP priors with shallow kernels (\eg RBF and Mat\'{e}rn) have achieved a great success in many applications, these shallow structures can limit the expressiveness in estimating highly complicated functions, \eg sharp discontinuities and high curvatures. To address this problem, \citet{wilson2016deep} proposed to construct deep kernels with neural networks. Specifically, they first choose a shallow kernel as the base kernel. Each input is first fed into a neural network (NN), and the NN outputs are then fed into the base kernel to compute the final kernel function value. Take RBF as an example of the base kernel, we can construct a deep kernel by
\begin{align}
k_{\text{DEEP}}(\x_i, \x_j) = k_{\text{RBF}}\left(\text{NN}(\x_i), \text{NN}(\x_j)\right). \label{eq:dk}
\end{align}
Note that the NN weights now become the kernel parameters. We can then use the deep kernel to construct a GP prior for nonparametric function estimation. The model likelihood and predictive distribution have the same forms as in \eqref{eq:gp} and \eqref{eq:predictive}, respectively. 

%DE form and settings; surrogate function --> sample a set of latent inputs to get projection --> raise the question about valid prior --> symmetric properties
%\vspace{-0.1in}
\section{Model} \label{sec:model}
%\vspace{-0.1in}
By using deep neural networks to construct highly expressive kernels, deep kernel learning greatly improves the capability of estimating complicated functions, and meanwhile inherits the self-adaptation of the nonparametric function learning and convenient posterior inference. However, as a purely data-drive approach, deep kernel learning can still suffer from data scarcity, especially when the training examples are inadequate to reflect the complexity of the underlying mapping and when the test points are distant  from all the training samples, \ie extrapolation. To overcome this limitation, we propose \ours, a physics informed deep kernel learning model that exploits physics prior knowledge to improve the function learning and uncertainty reasoning. Our model is presented as follows. 
\begin{figure*}[htbp]
	\centering
	\includegraphics[width=0.8\textwidth]{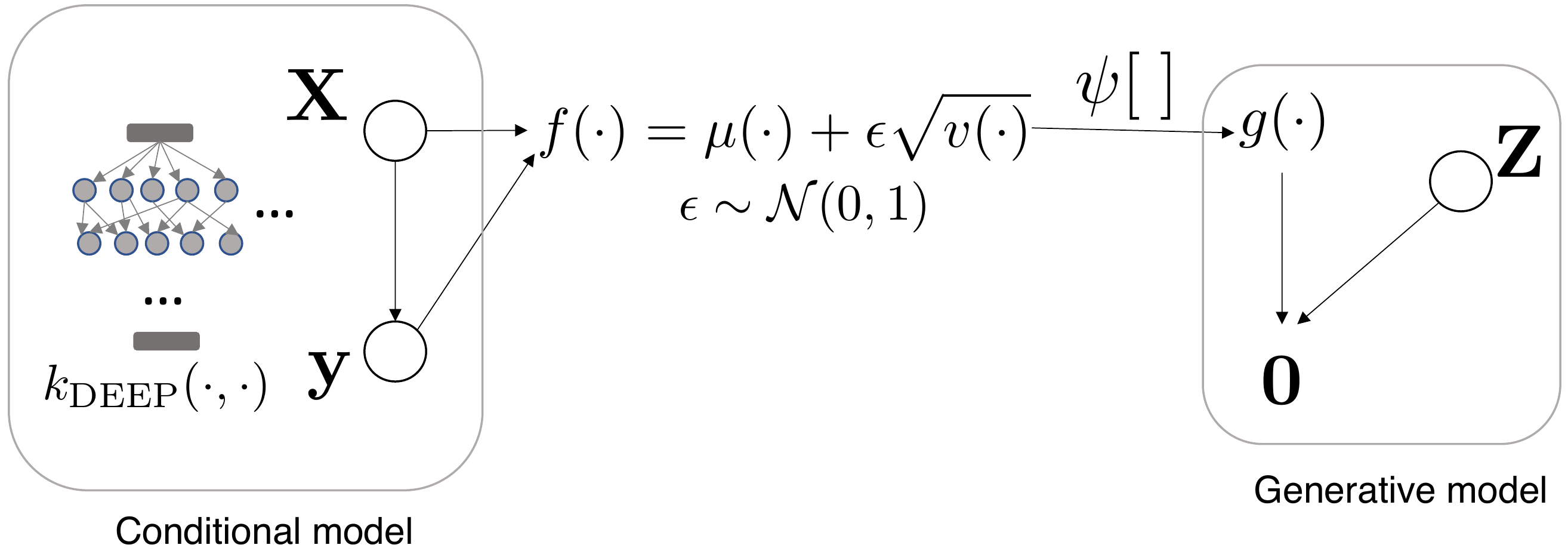}
	\caption{\small Graphical representation of Physics Informed Deep Kernel Learning (PI-DKL). The physics knowledge is encoded by a differential equation $\psi[f(\x)] = g(\x)$ where $g(\cdot)$ is an unknown source term. Note that $\x$ can include both the time and spatial variables. The conditional model is the deep-kernel GP and the generative model is equivalent to placing a GP prior over $g(\cdot)$.} \label{fig:graphical}
	\vspace{-0.1in}
\end{figure*}
%equation, form. 
%\vspace{-0.05in}
\subsection{Physics Informed Deep Kernel Learning}
%\vspace{-0.1in}
We assume that in general, the physics is  described by a  differential equation of the following form,
\begin{align}
\psi [f(\x)] = g(\x) \label{eq:pde} 
\end{align}
where $\psi$ is a functional that combines a set of differential operators, $f(\x)$ is the target (or solution) function we want to estimate from the training dataset $\Dcal = (\X, \y)$, and $g(\x)$ is a  \textit{latent source whose form is unknown}.  Note that the input of the latent source $g$ is in general assumed to be the \textit{same} as the solution $f$, \eg including both the \textit{spatial} and \textit{temporal} variables. The functional $\psi[\cdot]$ may include unknown parameters. One example is  $\psi[f(x)] = \frac{\d f(x)}{\d x} + \alpha f(x) - \beta$, where the input $x$ is a scalar, and $\alpha$ and $\beta$ are unknown parameters. This functional represents a linear operator. Another commonly seen example is from the viscous version of Burger's equation~\citep{olsen2011numerical}, 
$\psi [f(\x)] = \frac{\partial f(\x)}{\partial x_1}  + f(\x)\frac{\partial f(\x)}{\partial x_2} - v \frac{\partial^2 f(\x)}{\partial x_{2}^2}$, 
where $\x = [x_1, x_2]^\top$, $x_1$ is the spatial variable, $x_2$ the time variable, and $v$ the unknown viscosity. This functional includes a  nonlinear operator,  $f(\x)\frac{\partial f(\x)}{\partial x_2}$.

To incorporate the physics knowledge in \eqref{eq:pde}, we propose a hybrid of conditional and generative models based on the general framework of \citet{lasserre2006principled}. The conditional component is the standard deep-kernel GP that given the training inputs $\X$, samples the (noisy) output observations $\y$, and the probability $p(\y|\X)$ is given in \eqref{eq:gp}.  The generative component fulfills another GP prior over the latent source $g(\cdot)$, but avoids the double prior problem to ensure a valid joint probabilistic model for posterior inference. Coupled with the differential operators, the generative component regularizes and guides the deep kernel learning  of  $f(\cdot)$.  The graphical illustration of \ours is shown in Fig. \ref{fig:graphical}.   

Specifically, to consider a GP prior over $g(\cdot)$, we first sample a finite set of input locations $\Z = [\z_1, \ldots, \z_m]^\top$ (we will discuss the choice of $p(\Z)$ later). Then the projection of $g(\cdot)$ on $\Z$ follows a multivariate Gaussian distribution,  
\begin{align}
p(\g|\Z) = \N(\g|\0, \bSigma), \label{eq:gp-g} %=\N(\0|\g, \bSigma)
\end{align}
where $\g =  [g(\z_1), \ldots, g(\z_m)]^\top$, $[\bSigma]_{ij} = \kappa(\z_i, \z_j)$ and $\kappa(\cdot, \cdot)$ is another kernel. 

Next, we link the GP model of the target $f(\cdot)$ to the latent source $g(\cdot)$ via the differential equation \eqref{eq:pde}.  Our key idea is that from the GP posterior distribution \eqref{eq:predictive}, we can construct a sample of the target function, $f(\cdot) = \mu(\cdot) + \epsilon \sqrt{v(\cdot)}$\footnote{In computational physics, this is viewed as a surrogate for the solution function of the differential equation. This posterior function is also used in GP-UCB~\citep{DBLP:conf/icml/SrinivasKKS10}, a widely used Bayesian optimization algorithm for acquisition function calculation/optimization.}, 
%\begin{align}
%f(\cdot) = \mu(\cdot) + \epsilon \sqrt{v(\cdot)},
%\end{align}
where $\epsilon \sim \N(\epsilon|0, 1)$, $\mu(\cdot)$ and $\sqrt{v(\cdot)}$ are the posterior mean and standard deviation functions. While $f(\cdot)$ is a random function (due to $\epsilon$), it has a closed form and we can apply the functional $\psi$ to obtain the sample of $g(\cdot)$, 
\begin{align}
	g(\cdot) = h(\cdot, \epsilon) = \psi[\mu(\cdot) + \epsilon \sqrt{v(\cdot)}]. \label{eq:g-sample}
\end{align}
Therefore, to sample $\g$ --- the values of $g(\cdot)$ on $\Z$, we can first sample a standard Gaussian white noise $\epsilon$, and then sample from 
\begin{align}
	p(\g|\epsilon, \X, \y) = \prod_{j=1}^m \delta\left(\tg_j - h(\z_j, \epsilon)\right), \label{eq:gp-pde}
\end{align} 
where $\tg_j = g(\z_j)$ is $j$-th element of $\g$, and $\delta(\cdot)$ is the Dirac delta prior. We can also view $\g$ as a transformation of the Gaussian noise $\epsilon$ and derive the marginal distribution $p(\g|\X, \y)$ (see the discussion in the supplementary material), which, however, is much more difficult to compute. 

Now, we want to tie the GP prior for $g(\cdot)$ in \eqref{eq:gp-g} to the samples $\g$ generated from the GP model of the target function $f(\cdot)$ through $\eqref{eq:gp-pde}$. In this way, the learning of $f(\cdot)$ can be guided or regularized by the differential equation \eqref{eq:pde}. However, directly multiplying  \eqref{eq:gp-g} and \eqref{eq:gp-pde} is problematic, because $\g$ will have double priors, which are invalid in probabilistic modeling ~\citep{bishop2006pattern,wainwright2008graphical}. Note that every valid Bayesian model is defined by a probabilistic sampling procedure --- when $\g$ has already been sampled (according to $f(\cdot)$), we cannot use a second prior distribution(\ie \eqref{eq:gp-g}) to sample $\g$ again\footnote{unless we can show that their product is a new valid distribution, \ie the integral over the support is one.}; in other words, it breaks the DAG structure in the graphical model representation. 
% and the sampling procedure is invalid  --- if $\g$ is sampled from  \eqref{eq:gp-g}, it cannot be sampled again from \eqref{eq:gp-pde}, and vice versa. 
To ensure our model is a valid probabilistic model for posterior inference, we utilize the symmetric property of the Gaussian distribution, 
\begin{align}
p(\g|\Z) = \N(\g|\0, \bSigma) =\N(\0|\g, \bSigma) = p(\0|\g, \Z). \label{eq:gen-comp} 
\end{align}
We can see that placing a (finite) GP prior over $g(\cdot)$ is equivalent to sampling a set of virtual  data points $\0$. Therefore, we can turn the GP prior of the latent source to a generative component that samples the virtual data (observations) $\0$.  From the computational perspective, they are equivalent. However, the sampling procedure now becomes valid --- we first sample $\g$ from \eqref{eq:gp-pde}, and then sample the data $\0$ from \eqref{eq:gen-comp}. Note that the virtual data $\0$ come from the zero-mean function of the GP prior of $g(\cdot)$. We can use different virtual observations by adopting a nonzero mean function.  

Finally, we combine the conditional model  and the generative model (see \eqref{eq:gp}, \eqref{eq:gp-pde} and \eqref{eq:gen-comp}) to obtain a joint probability distribution,
\begin{align}
	&p(\y, \0, \Z,  \epsilon, \g|\X)  \notag \\
	& = p(\y|\X)p(\Z)p(\epsilon)p(\g|\epsilon, \X, \y) p(\0|\g, \Z) \notag \\
	&= \N(\y|\0, \K + \tau^{-1}\I) p(\Z) \N(\epsilon|0, 1) \notag \\  & \cdot\prod\nolimits_{j=1}^m \delta\left(\tg_j - h(\z_j, \epsilon)\right)  \N(\0|\g, \bSigma). \label{eq:joint}
\end{align} 
The choice of  $p(\Z)$ is flexible. If we have no knowledge about the input distribution, we can use a uniform distribution for the bounded domain, and for unbounded domains we can use  a wide Gaussian distribution with zero mean or uniform distribution on a region large enough to cover our interested inputs.

\vspace{-0.1in}
\section{Algorithm} \label{sec:algorithm}
\vspace{-0.1in}
\subsection{Stochastic Collapsed Inference}
\vspace{-0.1in}
We now present the model estimation algorithm. 
The exact posterior of the latent random variables $\Z$, $\epsilon$, and $\g$ in \eqref{eq:joint} are infeasible to calculate because they are coupled in kernels and differential operators. 
While we can use variational approximations, they will introduce extra variational parameters,  complicate the optimization and affect the integration of the physics knowledge. %This will also bring in more computational cost.  
Therefore, we marginalize out all the latent variables to conduct a collapsed inference to avoid approximating their complex posteriors. Specifically, we observe that $p(\y, \0|\X) = p(\y|\X)p(\0|\y, \X)$, where
\begin{align}
p(\0|\y, \X) = &\int p(\Z) p(\epsilon) p(\g|\epsilon,  \X, \y) p(\0|\g, \Z)  \d \Z \d \epsilon \d \g \notag \\
 =&\EE_{p(\Z)}\EE_{p(\epsilon)} [\int \delta(\g - \h) \N(\0|\g, \bSigma)  \d \g] \notag \\ = &\EE_{p(\Z)}\EE_{\N(\epsilon|0, 1)} \left[\N\left(\h(\Z, \epsilon)|\0, \bSigma\right)\right]. \label{eq:gen_marginal}
\end{align}
where $\h(\Z, \epsilon) = [h(\z_1, \epsilon), \ldots, h(\z_m, \epsilon)]^\top$. Note that $h(\cdot, \cdot)$ is defined in \eqref{eq:g-sample}. Thanks to the automatic differentiation techniques, we can  directly evaluate the differential operators in $\psi$ over $f(\cdot)$ in \eqref{eq:g-sample}. We never need to use meshes, discretization, \etc as in numerical solvers.
% This also strengthens the physics knowledge integration --- when we integrate out the input locations $\Z$, we essentially impose the probability constraints (see \eqref{eq:zeros}) at all possible sets of $m$ input locations, rather than just one. 

To allow us to flexibly adjust the \cmt{importance of the generative component and so} the influence of the physics during training, we introduce a weighted likelihood of the generative component~\citep{warm1989weighted,hu2002weighted} by a free hyper-parameter $\gamma \ge 0$. The weighted marginal likelihood~\citep{warm1989weighted,hu2002weighted} is given by
\begin{align}
p_\gamma (\y, \0|\X) = p(\y|\X) p(\0|\X, \y)^{\gamma}.
\end{align} 
Our inference is to maximize the log weighted marginal likelihood to optimize the kernel parameters in $k_{\text{DEEP}}(\cdot, \cdot)$ and $\kappa(\cdot, \cdot)$, the inverse noise variance $\tau$ and unknown parameters in the differential equation, $\log p_\gamma (\y, \0|\X) =  \log \big(\N(\y|\0, \K + \tau^{-1}\I)\big) + \gamma \log \big(\EE_{p(\Z)}\EE_{\N(\epsilon|0, 1)} \left[\N\left(\h(\Z, \epsilon)|\0, \bSigma\right)\right]\big)$. This log likelihood is infeasible to compute due to the intractable expectation inside the logarithm. To address this issue, we use Jensen's inequality on the log function to obtain an evidence lower bound (ELBO),  $\Lcal \le \log p_\gamma (\y, \0|\X) $ where 
\begin{align}
\Lcal = &  \log \big(\N(\y|\0, \K + \tau^{-1}\I)\big)  \notag \\ + & \gamma \cdot \EE_{p(\Z)}\EE_{\N(\epsilon|0,1)} \left[\log\left(\N\big(\h(\Z, \epsilon)|\0, \bSigma\big)\right)\right]. \label{eq: elbo}
\end{align}
While $\Lcal$ is still intractable, it is straightforward to maximize $\Lcal$ with stochastic optimization. Each time, we generate a sample of  the input locations from $p(\Z)$ and the noise from $\N(\epsilon|0, 1)$, denoted by $\tZ$ and $\teps$.  We then obtain $\widetilde{\Lcal} = \log \big(\N(\y|\0, \K + \tau^{-1}\I)\big)  +  \gamma  \log\left(\N\big(\h(\tZ, \teps)|\0, \bSigma\big)\right)$,  an unbiased estimation of $\Lcal$. We calculate $\nabla \widetilde{\Lcal}$ as an unbiased stochastic gradient of $\Lcal$, with which we can use any stochastic optimization to estimate the model parameters. While $\h(\cdot, \cdot)$ couples the deep kernels and complex operators in $\psi$, it is differentiable and we can use automatic differentiation libraries to calculate the stochastic gradient. 

The ELBO $\Lcal$ in \eqref{eq: elbo} is  the GP log marginal likelihood plus an extra term, $\EE_{p(\Z)}\EE_{\N(\epsilon|0,1)} \left[\log\big(\N\big(\h(\Z, \epsilon)|\0, \bSigma\big)\big)\right]$. {Each element of $\h$ is obtained by applying the functional $\psi$   on the posterior sample of  $f(\cdot)$ (see \eqref{eq:g-sample}).} Jointly maximizing this term in $\Lcal$ encourages that the latent source values (at $m$ locations) obtained from the GP posterior function $f(\cdot)$ (through the equation) should be considered as the samples of another GP.  This can be viewed as a soft constraint over the posterior function of the GP. Therefore, our ELBO is also a posterior regularization objective~\citep{ganchev2010posterior}, and our inference  algorithm estimates the standard deep-kernel GP model with a soft regularization on its posterior distribution.

\vspace{-0.1in}
\subsection{Algorithm Complexity}
\vspace{-0.1in}
The time complexity for the inference of our model is $\Ocal(N^3 + m^3)$, because it involves the calculation for two GPs: one is the standard GP, and the other is reflected in the generative component. The time complexity for prediction is still $\Ocal(N^3)$. The space complexity is $\Ocal(N^2 + m^2)$, including the storage of the kernel matrices of the two GPs.

%linear PDE and GP , LFM
%George's work?
%posterior regularization 
%\vspace{-0.1in}
\section{Related Work}
\vspace{-0.1in}
%discuss about PINN; difference with our work and scalabl version
An influential work, physics informed neural networks (PINNs)~
\citep{raissi2019physics}, was recently proposed to train neural networks that respect physical laws. The key idea is to use neural networks as a surrogate for the solution of the (partial) differential equation, and minimize the NN loss plus the residual error of the equation on a set of randomly collected collocation points in the input domain. The follow-up research includes ~\citep{mao2020physics,jagtap2020adaptive,zhang2020learning,chen2020physics,pang2019fpinns}, \etc While our work is enlightened by PINNs, there are several substantial differences. First, PINNs demand the form of the PDE is fully specified, \ie $\psi[f(\x)]  = 0$,  while we assume that there can be some unknown source function, $\psi[f(\x)] = g(\x)$. Thus, our work is to exploit  incomplete physics knowledge. Second, we use the posterior of the deep-kernel GP to construct a random surrogate for the PDE solution, and cast the integration of the physics into a  principled Bayesian framework to enable posterior inference and uncertainty quantification, while PINNs only conduct point estimations. Our experiments show that the incomplete physics knowledge can also improve the uncertainty quantification. Note that \citet{zhang2019quantifying} have combined polynomial chaos~\citep{xiu2002wiener} and dropout~\citep{gal2016dropout} to estimate the total uncertainty for PINNs with stochastic PDEs.

Many works have used GPs to model or learn physical systems~\citep{graepel2003solving,lawrence2007modelling,gao2008gaussian, alvarez2009latent,alvarez2013linear, raissi2017machine}. For example, \citet{graepel2003solving} uses GPs to solve the linear equation given observed noisy sources only. \cmt{He first defines the kernel for the solution function with which to derive the kernel for the source function. The kernel parameters are then estimated from the noisy source data, given which the solution can be predicted.} \citet{raissi2017machine} assume both the noisy sources and solutions are observerable, and they jointly model these examples in one single GP with a heterogeneous block covariance matrix. Other works include~\citep{calderhead2009accelerating,barber2014gaussian,macdonald2015controversy, heinonen2018learning,lorenzi2018constraining,wenk2019fast,wenk2020odin,pan2020physics} \etc They mainly focus on estimating parameters/operators in Ordinary Differential Equations (ODEs) and do not consider latent sources. Although the excellent work of \citet{lorenzi2018constraining} can incorporate general equality or inequality constraints based on both ODEs and PDEs, it strictly requires every function in the constraint has an explicit form. A critical difference is that \ours aims to incorporate \textit{arbitrary}, \textit{incomplete} differential equations to improve deep kernel learning on scarce data, in both prediction accuracy (especially extrapolation) and uncertainty quantification.

%Latent force models (LFM)~\citep{alvarez2009latent} make the same assumption about the differential equations as our work. LFMs convolve the Green's function of the equation with the kernel for the latent source to obtain the kernel for the target, and then learn the kernel parameters from data. While LFMs enable a hard encoding of physics, they rely on an analytical Green's function, which is not available for many equations. In addition, LFMs construct shallow kernels, which can be less expressive as deep kernels.  
The classical Latent force models (LFM)~\citep{alvarez2009latent} also assume unknown, unobserved source terms in the equation.  LFMs convolve the Green's function of the equation with the kernel for the latent sources to obtain the kernel for the target, and then learn the kernel parameters from data.  Hence, the original LFMs rely on an analytical Green's function, which might not be available for many (nonlinear) equations. The recent excellent works~\citep{hartikainen2012state,ward2020black} have overcome this limitation when the source terms are functions of \textit{time} only. They construct a companion stochastic differential equation (SDE) and infer the hidden state (including the solution, source terms, their time derivatives, \etc) via non-linear Kalman filtering~\citep{hartikainen2012state} or black-box variational inference~\citep{ward2020black}. Note that our work focuses on general latent sources, \ie functions of \textit{all} the inputs, including both the \textit{time} and \textit{spatial} variables. In this case, these approaches can still be difficult to apply.  See Sec \ref{exp:pm25} and \ref{exp:pm25} for detailed examples. 
%While LFMs enable a hard encoding of physics, they rely on an analytical Green's function, which is not available for many equations. In addition, LFMs construct shallow kernels, which can be less expressive as deep kernels.  

%It is known that applying a linear (partial) differential operator on a GP will result in another GP~\citep{graepel2003solving}. Many excellent works have been done in this direction~\citep{graepel2003solving,lawrence2007modelling,gao2008gaussian, alvarez2009latent,alvarez2013linear, raissi2017machine}. For example, \citet{graepel2003solving} uses GPs to solve the linear equation given observed noisy forces ($u_q(\cdot)$ in \eqref{eq:lineq}). He first defines the kernel for the solution function ($f_q(\cdot)$ in \eqref{eq:lineq}) with which to derive the kernel for the forces. The kernel parameters are then estimated from the noisy forces data, given which the solution can be predicted. \citet{raissi2017machine} assume both the noisy forces and solutions are observed, and they jointly model these examples in one single GP with a heterogeneous block covariance matrix. Other excellent works related to GP derivatives include~\citep{calderhead2009accelerating,barber2014gaussian, heinonen2018learning} \etc They mainly focus on estimating parameters/operators in ODEs without latent functions/forces as assumed in LFM and our work.% (see \eqref{eq:lineq} and \eqref{eq:pde}). 

\cmt{
Posterior regularization is a powerful inference methodology~\citep{ganchev2010posterior}. In general, the objective includes the model likelihood on data and a penalty term that encodes the constrains over the posterior of the latent variables. %Via the penalty term, we can incorporate our domain knowledge or constraints outright to the posteriors, rather than through the priors and a complex, intermediate computing procedure, hence it can be more convenient and effective. 
Many successful posterior regularization algorithms have been proposed, \eg ~\citep{he2013graph,ganchev2013cross,zhu2014bayesian,bilen2014weakly,libbrecht2015entropic,song2016kernel}.
While our inference algorithm is developed for a Bayesian hybrid model,  the ELBO optimized in the inference is a typical posterior regularization objective  that estimates  a standard GP model and meanwhile penalizes the posterior of the (target) function to encourage the consistency with the differential equations. This aligns with our modeling goal. 
}
\cmt{
%\vspace{-0.1in}
It is known that applying a linear (partial) differential operator on one GP will result in another GP~\citep{graepel2003solving}. This can be seen based on the fact that the derivative of GP is still a GP~\citep{williams2006gaussian}.
The covariance of a GP's derivative and the cross-covariance between the GP and its derivative can be obtained by taking derivatives over the original covariance (or kernel) function.  Hence, we can easily make a connection between the (latent) forces and solutions in linear equations (see \eqref{eq:lineq}) in the GP framework. Many excellent works have been done in this direction~\citep{graepel2003solving,lawrence2007modelling,gao2008gaussian, alvarez2009latent,alvarez2013linear, raissi2017machine}. For example, \citet{graepel2003solving} uses GPs to solve the linear equation given observed noisy forces ($u_q(\cdot)$ in \eqref{eq:lineq}). He first defines the kernel for the solution function ($f_q(\cdot)$ in \eqref{eq:lineq}) with which to derive the kernel for the forces. The kernel parameters are then estimated from the noisy forces data, given which the solution can be predicted. \citet{raissi2017machine} assume both the noisy forces and solutions are observed, and they jointly model these examples in one single GP that includes a heterogeneous block covariance matrix.  However, in many applications, the force terms are unknown and not observable. To address this problem,  \citet{alvarez2009latent, alvarez2013linear} proposed the latent force model (LFM), which only uses the examples of the noisy solutions values. LFM first places a prior over the latent forces, and then derives the covariance (kernel)  of the solution function via the convolution operation. Despite the success of LFM in multi-output (or multitask) regression applications, such as transcriptional regulation modeling~\citep{lawrence2007modelling} and inference  of latent chemical species~\citep{gao2008gaussian}, the requirement of linear  equations and analytical Green's functions might be too restrictive, and prevent us from exploiting nonlinear equations or equations without tractable Green's functions. In addition, the convolution procedure (\ie integral) might restrict the usage of complex yet expressive kernels. 
To overcome these limitations, we propose physics regularized GP (PRGP), which starts with a (highly) flexible/expressive kernel for the solution function, and then use another GP for the latent forces to regularize the original GP through the differential operations. The regularization is fulfilled via a valid generative model component rather than process differentiation, and hence can be applied to any linear/nonlinear differential operators. Our approach also supports the adjustment (tuning) of importance for the physics. Compared with LFM, a hard-encoding approach, our method conducts soft-encoding and is much more flexible in exploiting different types of differential equations and expressive kernels, and allow us to tune the influence of the physics in training. %This can be particularly useful 

Other excellent works related to GP derivatives include~\citep{calderhead2009accelerating,barber2014gaussian, heinonen2018learning} \etc They mainly focus on estimating parameters/operators in ODEs without latent functions/forces as assumed in LFM and our work (see \eqref{eq:lineq} and \eqref{eq:pde}). 

Posterior regularization is a powerful inference methodology~\citep{ganchev2010posterior}. In general, the objective includes the model likelihood on data and a penalty term that encodes the constrains over the posterior of the latent variables. Via the penalty term, we can incorporate our domain knowledge or constraints outright to the posteriors, rather than through the priors and a complex, intermediate computing procedure, hence it can be more convenient and effective. 
A variety of successful posterior regularization algorithms have been proposed, \eg ~\citep{he2013graph,ganchev2013cross,zhu2014bayesian,bilen2014weakly,libbrecht2015entropic,song2016kernel}.
While our inference algorithm is developed for a new model (rather than pure GP),  the evidence lower bound optimized by our algorithm is a typical posterior regularization objective  that estimates  a pure GP model and meanwhile penalizes the posterior of the (target) function to encourage a consistency with the differential equations. This also aligns with our modeling goal. 
}

%physics informed NNs, kriging 
%hybrid modeling , NaNOS, eigen-net
%small data, visualization large data
%\vspace{-0.15in}
\section{Experiments} \label{sec:experiments}
\vspace{-0.05in}
\subsection{Simulation}
\vspace{-0.05in}
We first examined if \ours can improve extrapolation with right physics knowledge. We generated two synthetic datasets. The first dataset, \textit{1stODE}, was simulated from a first-order ODE, %$\frac{\partial f(t)}{\partial t} + B\cdot  f(t) - D = g(t)$
  \begin{align}
  \frac{\partial f(t)}{\partial t} + B\cdot  f(t) - D = g(t), \label{eq:ode}
  \end{align}
  where $B = D = 1$, $g(t) = \sin(2\pi t)\exp(-t)$ and the initial condition $f(0) = 0.1$. We set the time domain $t \in [0, 1]$. We ran the finite difference algorithm~\citep{mitchell1980finite} to obtain the accurate solution. We chose $1,001$ equally spaced time points ($t_0 = 0, t_{1000}  = 1$) and their solution values as the dataset. %The first and last time points are $0$ and $1$, respectively, and so the space is $0.001$.  %We used the first $101$ data points, \ie $t \in [0, 0.1]$,  for training, and the remaining $900$ data points, \ie $t \in (0.1, 1]$, for test. 
  The second dataset, \textit{1dDiffusion}, was simulated from a diffusion equation with one dimensional spatial domain, %$\frac{\partial f(x, t)}{\partial t} - \alpha \frac{\partial f^2(x,t)}{\partial x^2} = g(x, t)$
  \begin{align}
  \frac{\partial f(x, t)}{\partial t} - \alpha \frac{\partial f^2(x,t)}{\partial x^2} = g(x, t), \label{eq:heat}
  \end{align}
  where $\alpha = 10$, $g(x, t) = 0$ and the initial condition $f(x, 0)$ is a square wave. \cmt{Note that we chose a constant latent function $g$ to be consistent with the setting of the LFM paper~\citep{alvarez2009latent}.}  We set the domain $(x,t) \in [0, 1] \times [0, 1]$. We ran a numerical solver to obtain the accurate solution. Then we discretized the entire spatial and time domain into a $48 \times 101$ grid with equal spacing in each dimension. We retrieved the grid points and their solution values as our dataset. 

	\noindent \textbf{Competing methods}. We compared with (1) shallow kernel learning (SKL) with SE-ARD kenrel, (2) deep kernel learning (DKL), and (3)  LFM, which uses SE-ARD for the latent source, and then convolves it with Green's function to obtain the kernel for the target function.  To construct a deep kernel, we followed~\citep{wilson2016deep} to feed the input variables to a (deep) neural network (NN) and calculated the RBF kernel over the neural network outputs (see \eqref{eq:dk}). Across our experiments, we used a 5-layer NN, with 20 nodes in each hidden and  output layer. We used $\mathrm{tanh}(\cdot)$	 as the activation function. For our method \ours, we used the same deep kernel for the  target function. As in LFM, we used SE-ARD kernel for the latent source. % for \textit{1stODE} and the delta kernel for \textit{1dDiffusion} (to account for the constant latent function). 
	We set the number of virtual observations $m=10$  for the generative component, and uniformly sampled the input locations from the entire domain (see \eqref{eq: elbo}). We chose the weight of the generative component  $\gamma$ from $\{0.01, 0.05, 0.1, 0.5, 1, 2, 5, 10\}$ via cross-validation on the training data.
	For both LFM and \ours, the parameters of differential equations are unknown.  All the methods were implemented with TensorFlow~\citep{abadi2016tensorflow}. For our method, we used ADAM~\citep{kingma2014adam} for stochastic inference. We ran 10K epochs to ensure convergence. For the other methods, we used L-BFGS for optimization and set the maximum number of iterations to 5K. 

\begin{figure*}
	\centering
	\setlength\tabcolsep{0pt}
	\begin{tabular}[c]{cccc}
		\begin{subfigure}[b]{0.25\textwidth}
			\centering
			\includegraphics[width=\linewidth]{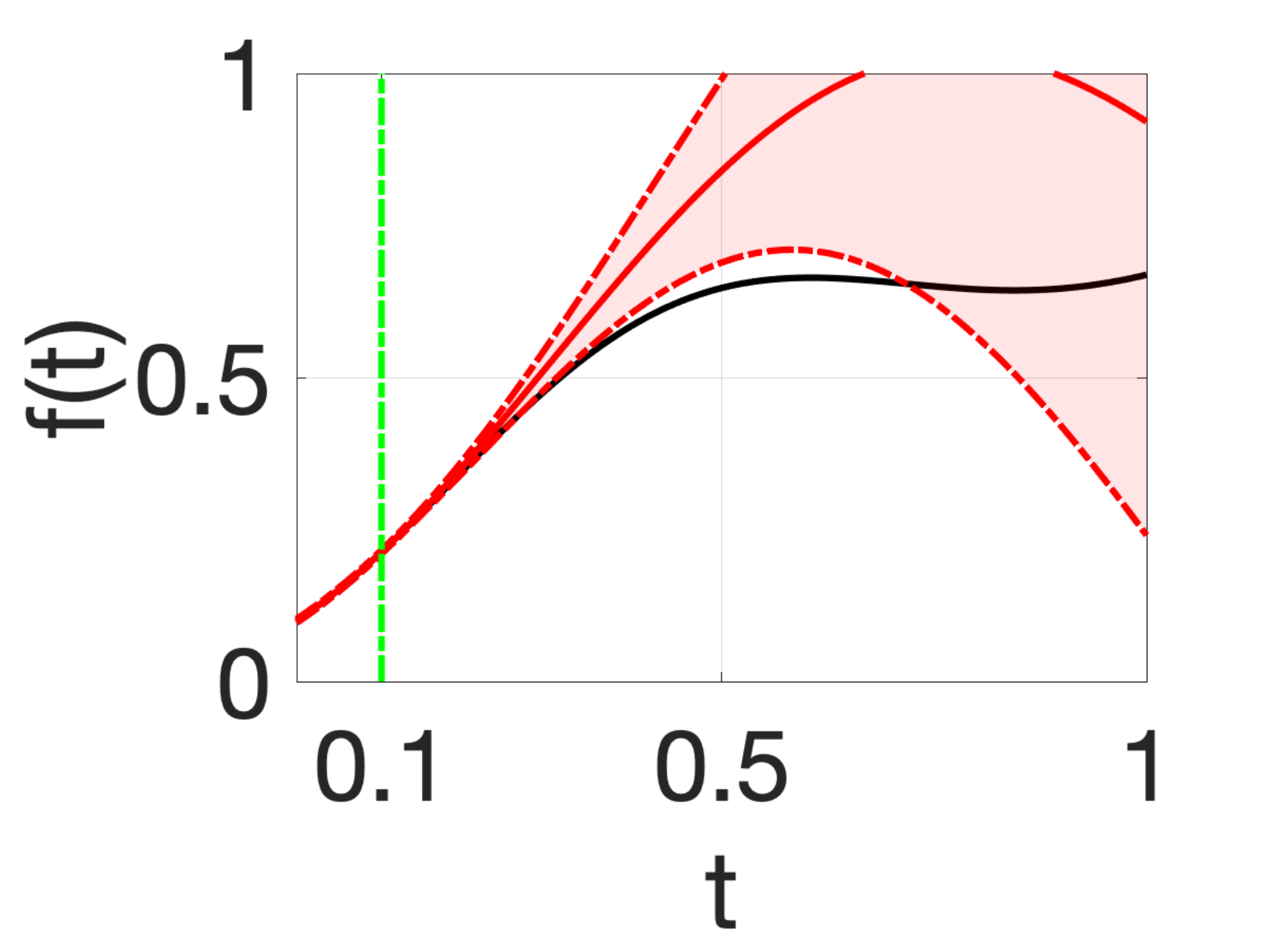}
			\caption{SKL}
		\end{subfigure}
		&
		\begin{subfigure}[b]{0.25\textwidth}
			\centering
			\includegraphics[width=\linewidth]{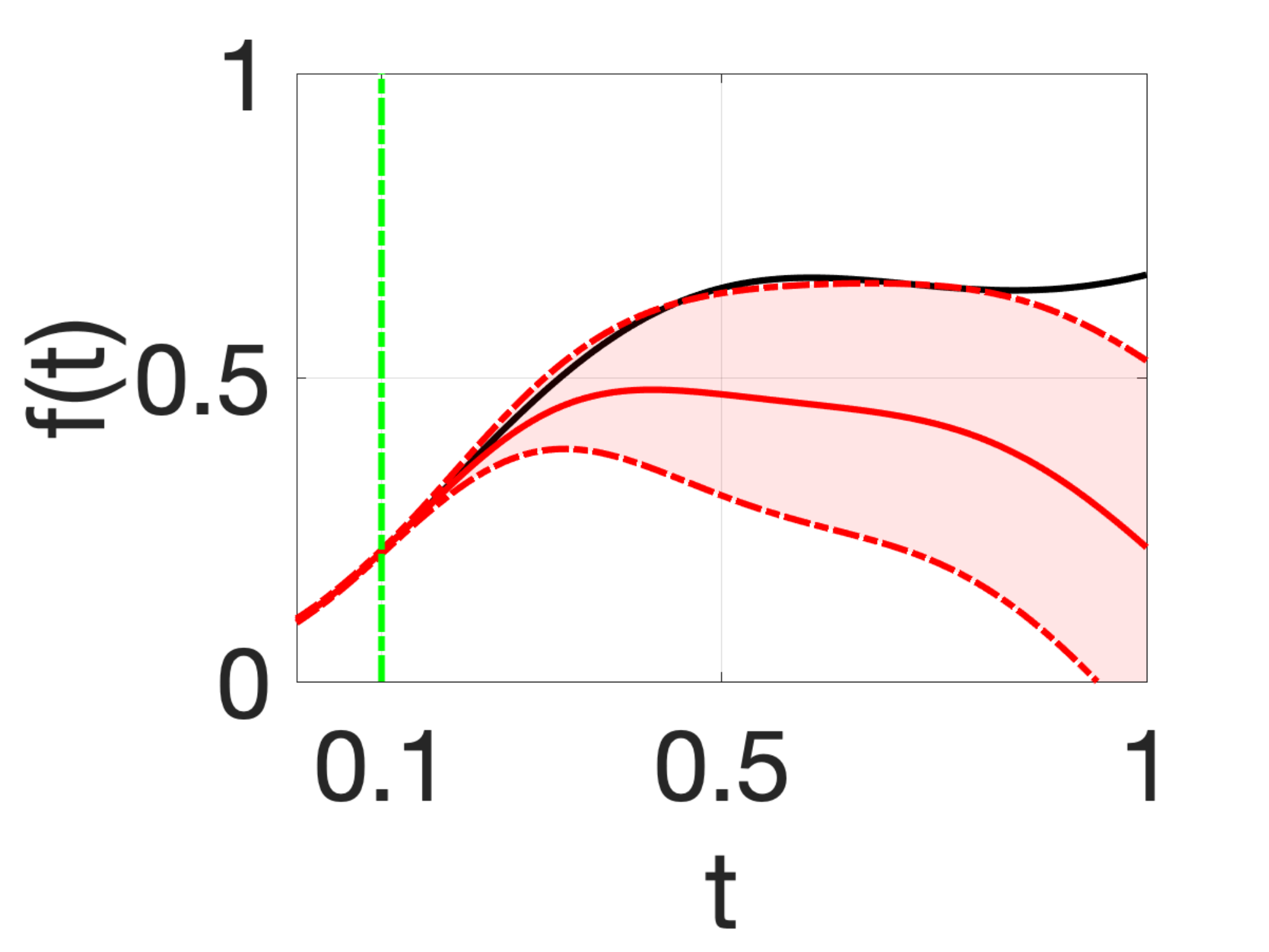}
			\caption{DKL} 
		\end{subfigure}
		&
		\begin{subfigure}[b]{0.25\textwidth}
			\centering
			\includegraphics[width=\linewidth]{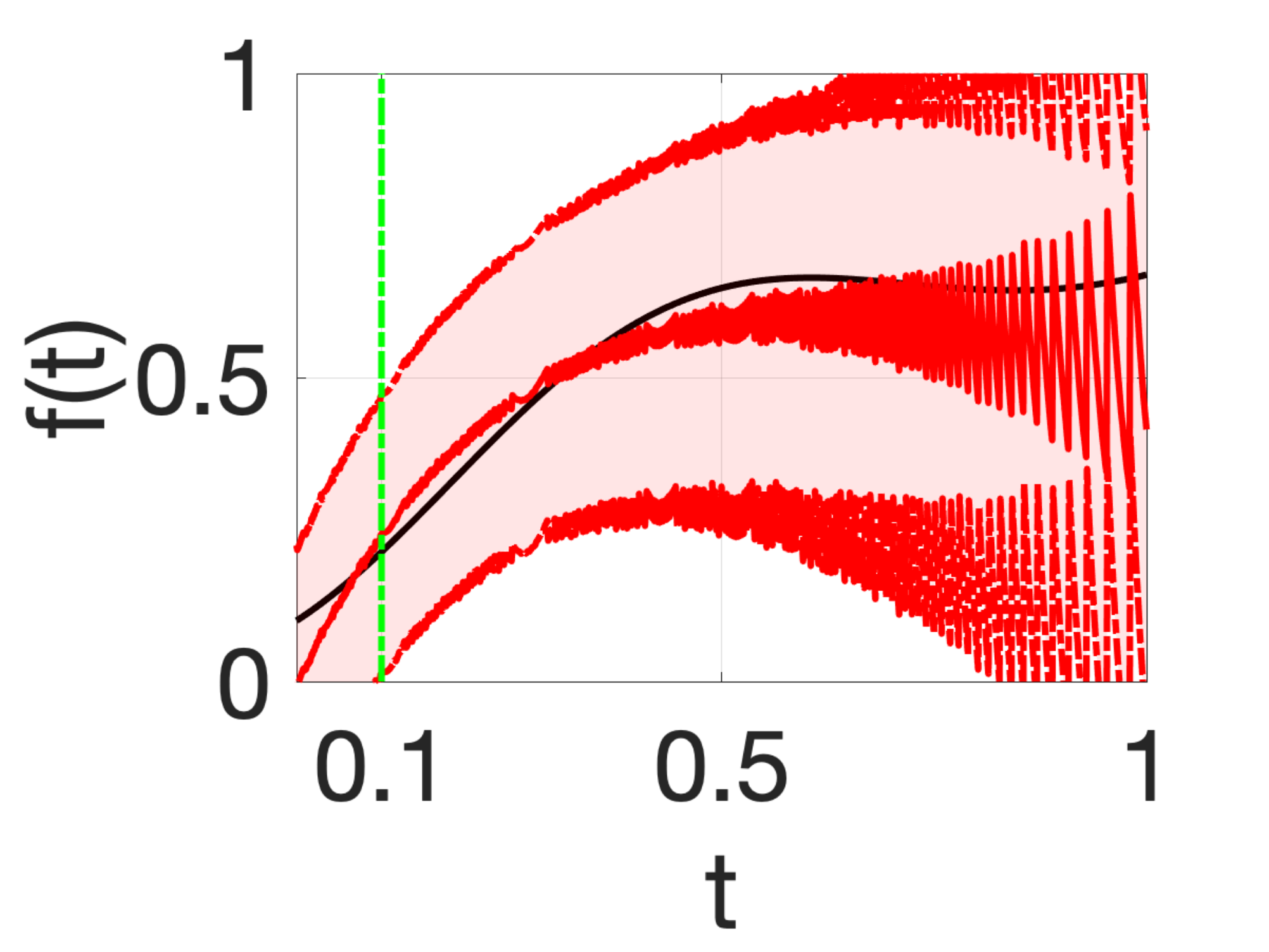}
			\caption{LFM} 
		\end{subfigure}
		&
		\begin{subfigure}[b]{0.25\textwidth}
			\centering
			\includegraphics[width=\linewidth]{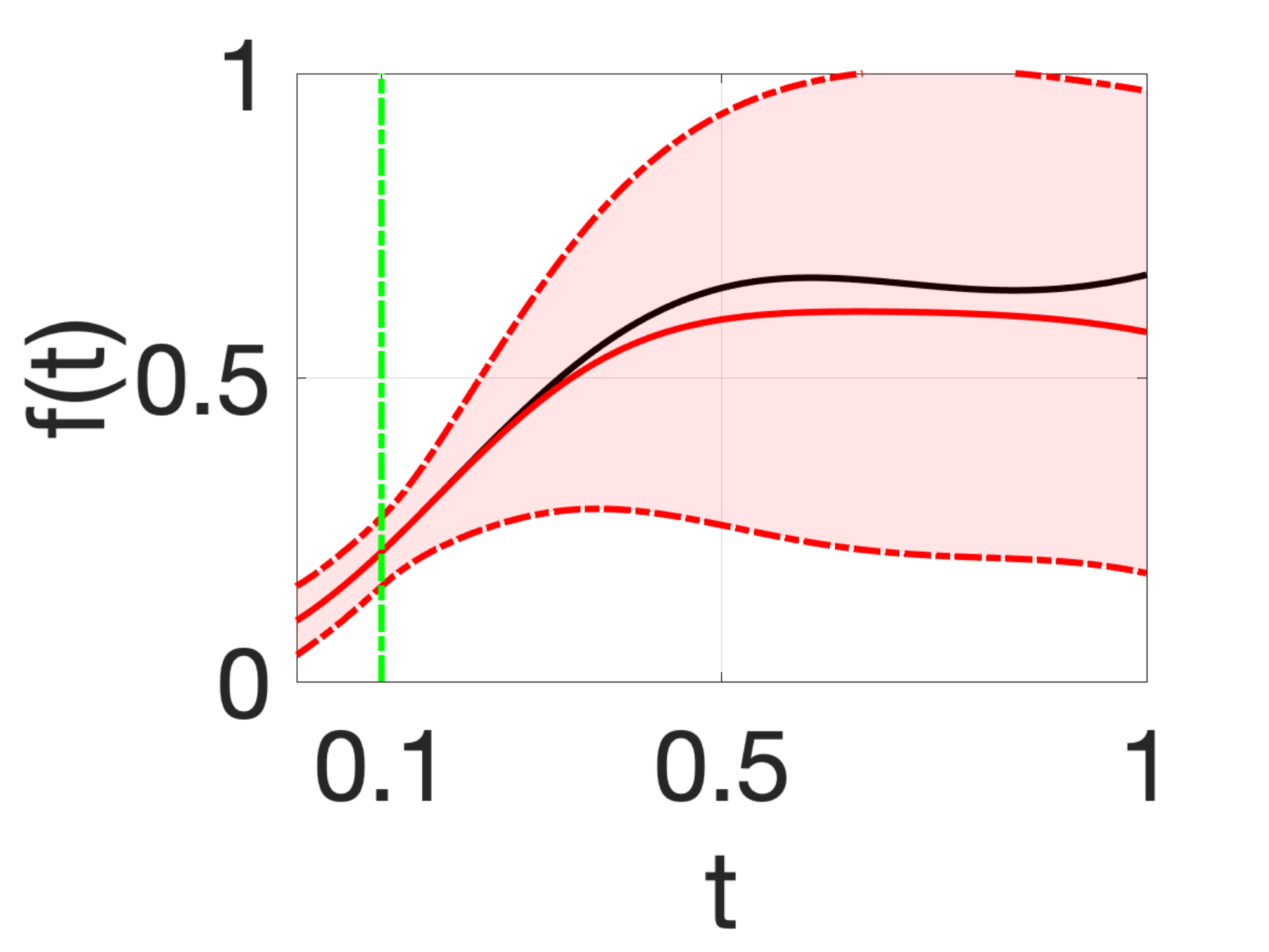} 
			\caption{\ours} 
		\end{subfigure}
	\end{tabular}
	\vspace{-0.15in}
	\caption{\small The posterior distribution of the learned solution functions on \textit{1stODE}. The red lines in the middle are the posterior means and the red dashed lines on the boundary of the shaded region the posterior mean plus/minus one posterior standard deviation. The black line is the ground-truth solution. The training inputs stay in $[0, 0.1]$ (left to the green line). }
	\vspace{-0.1in}
	\label{fig:ode}
\end{figure*}

 For \textit{1stODE} , we used the first $101$ samples ($t_i \in [0, 0.1]$) for training, and the remaining $900$ samples ($t_i \in (0.1, 1]$) for test. We show the posterior distribution of the functions learned by all the methods and the ground-truth in Fig. \ref{fig:ode}. We can see that the predictions of SKL and DKL are largely biased when the test points are far from the training region $[0, 0.1]$. On average, DKL obtains better accuracy than SKL. The root-mean-square errors (RMSEs) are \{DKL:0.21, SKL:0.25\}. As a comparison, the posterior means of LFM and \ours are much closer to the ground-truth in the test region, and the RMSEs are \{LFM: 0.09, \ours: 0.04\}, showing the benefit of the physics. However, LFM is quite unstable in extrapolation: the farther away the test area, the more fluctuating the prediction. By contrast, \ours obtains much smoother curves that are even closer to the ground-truth,  and smaller posterior variances in the test region. Hence, it shows that the LFM kernel obtained from the shallow kernel convolution is less expressive/powerful than the regularized deep kernel in \ours.  Note that unlike SKL/DKL, both LFM and \ours estimated nontrivial posterior variances (\ie not extremely close to 0) in the training region, implying that the physics  also helps prevent over-fitting.% the training data. 

\begin{figure*}
	\vspace{-0.45in}
	\centering
	\setlength\tabcolsep{0pt}
	\renewcommand{\arraystretch}{0}
	\begin{tabular}[c]{ccccc}
		\begin{subfigure}[b]{0.05\textwidth}
			\raisebox{0.4in}
			{
				\includegraphics[scale=0.15]{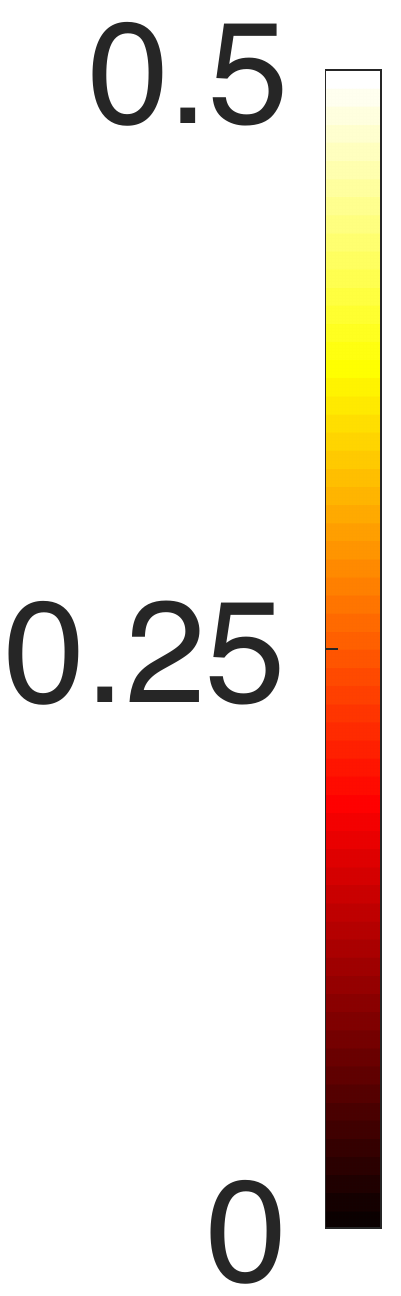}
			}
		\end{subfigure}
		\setcounter{subfigure}{0}
		&
		\begin{subfigure}[b]{0.24\textwidth}
			\centering
			\includegraphics[width=\linewidth]{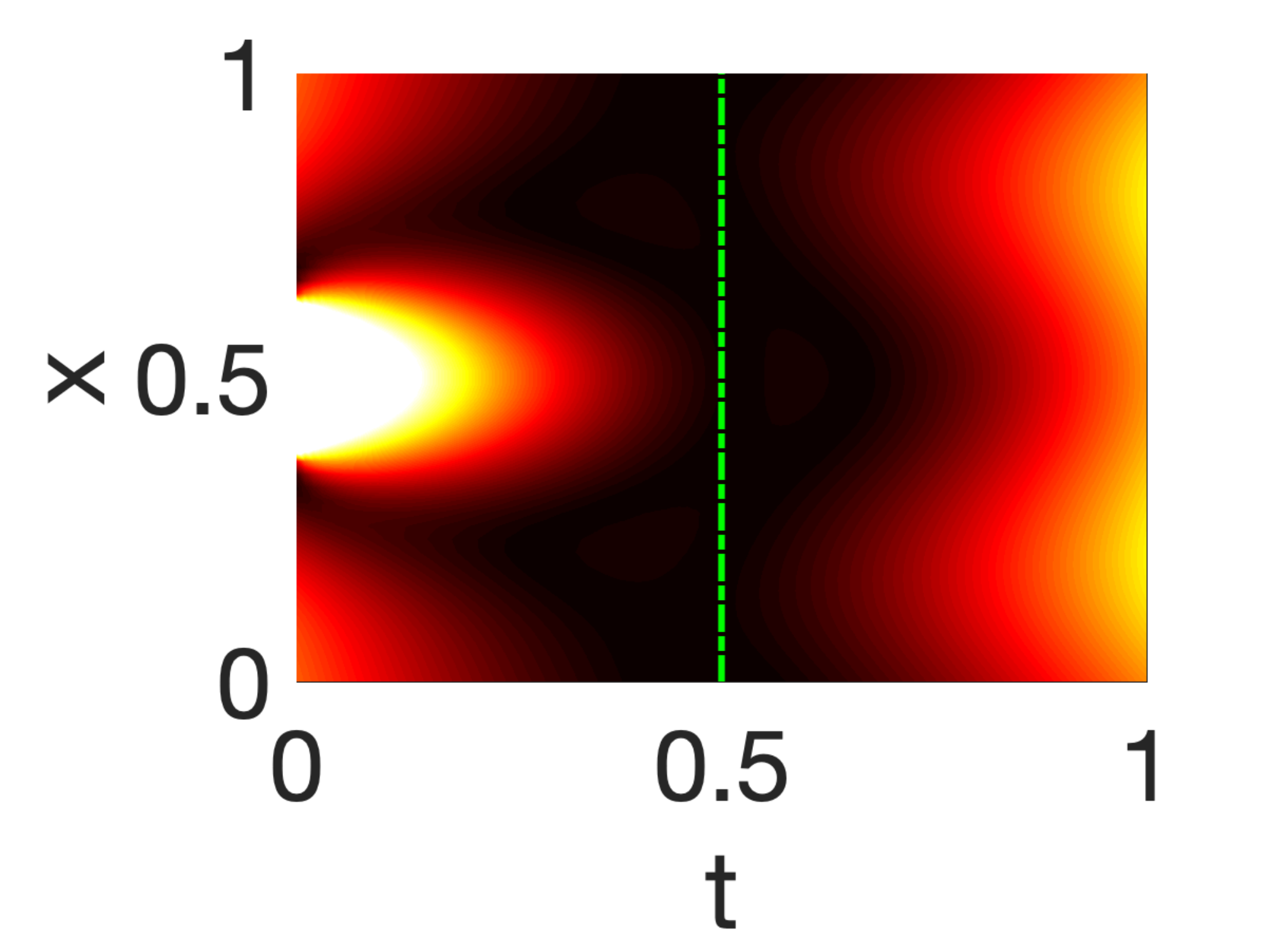}
			\caption{\small SKL } 
		\end{subfigure}
		&
		\begin{subfigure}[b]{0.24\textwidth}
			\centering
			\includegraphics[width=\linewidth]{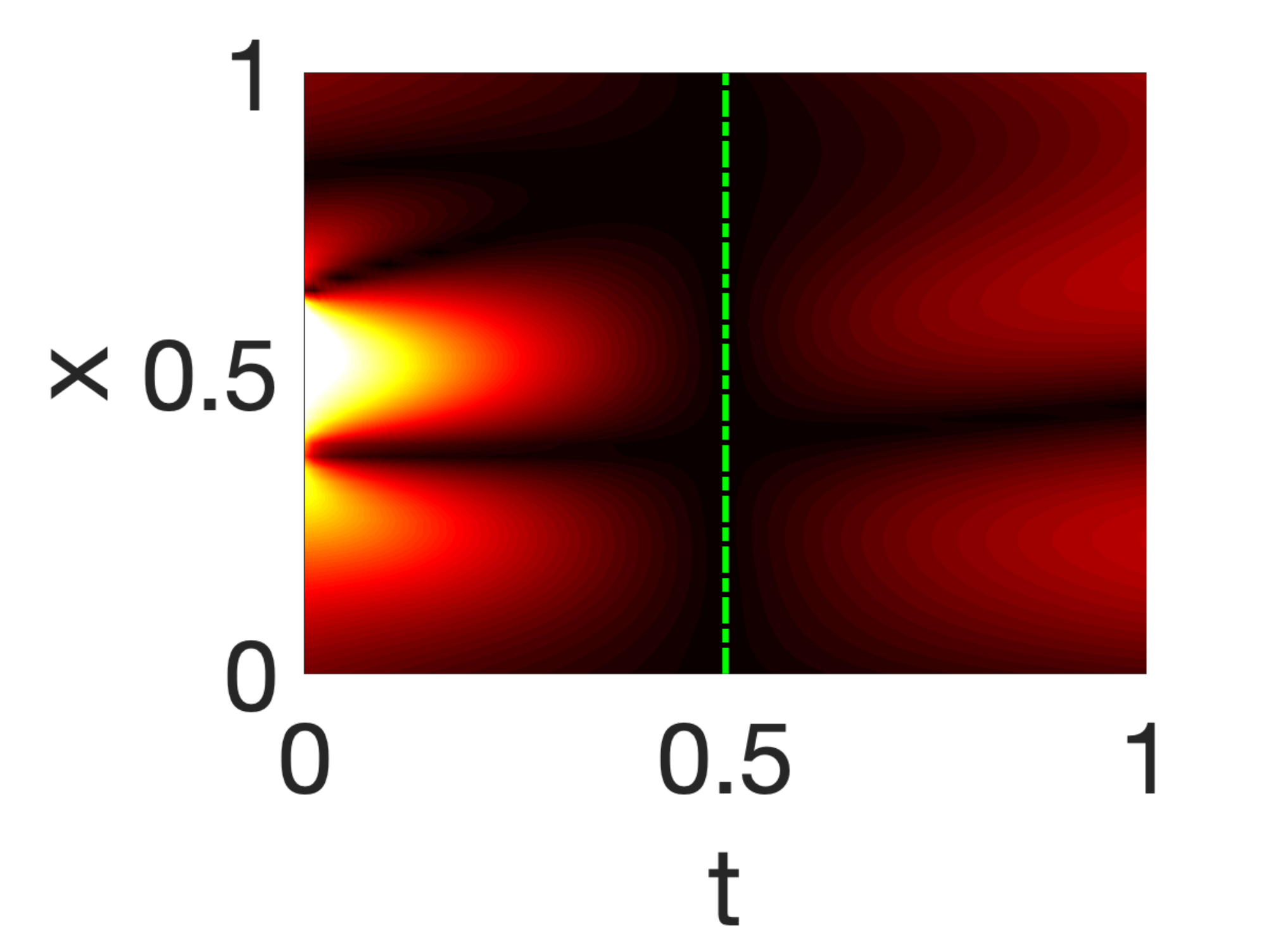}
			\caption{\small DKL } 
		\end{subfigure}
		&
		\begin{subfigure}[b]{0.24\textwidth}
			\centering
			\includegraphics[width=\linewidth]{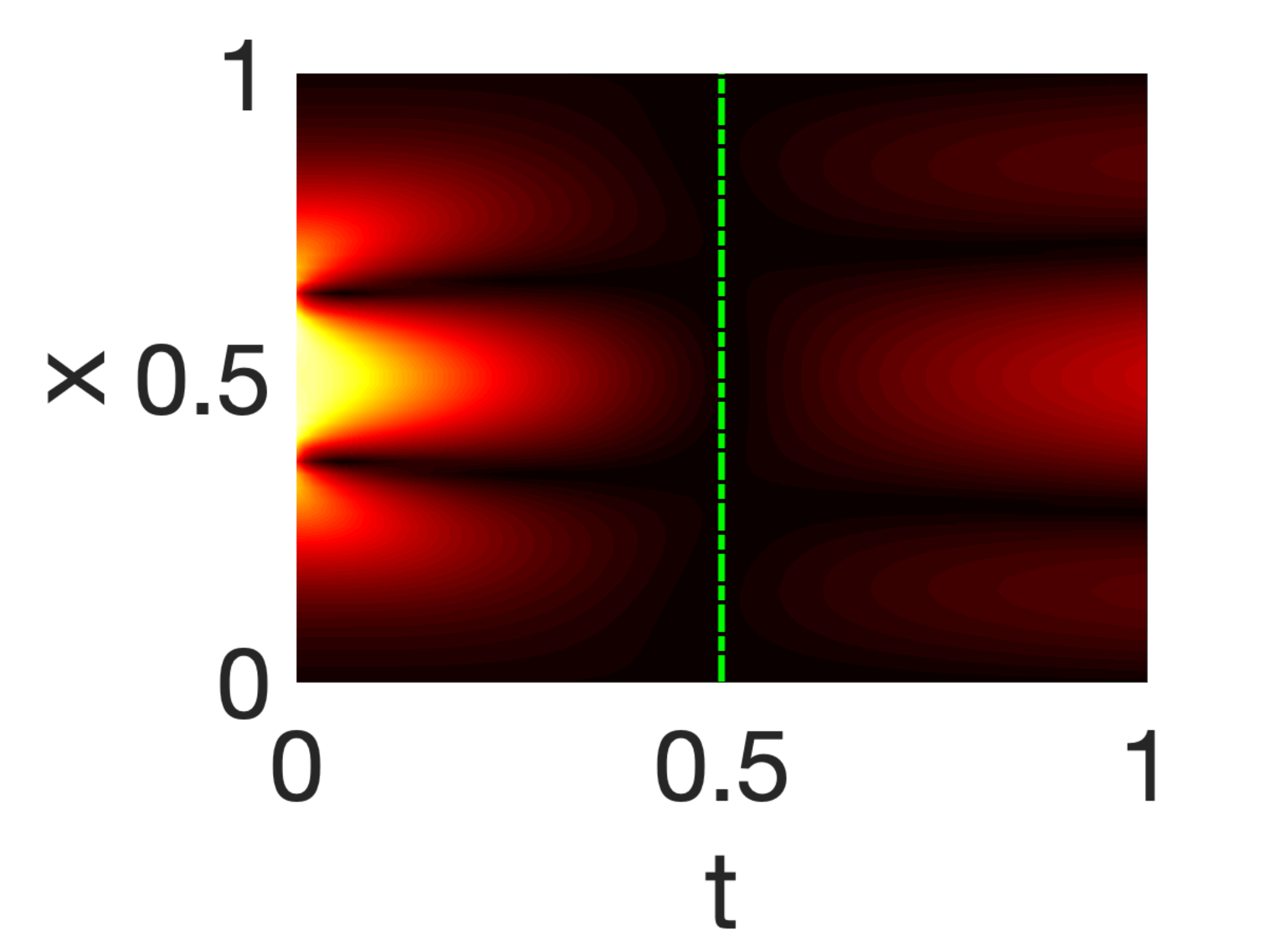}
			\caption{\small LFM} 
		\end{subfigure}
		&
		\begin{subfigure}[b]{0.24\textwidth}
			\centering
			\includegraphics[width=\linewidth]{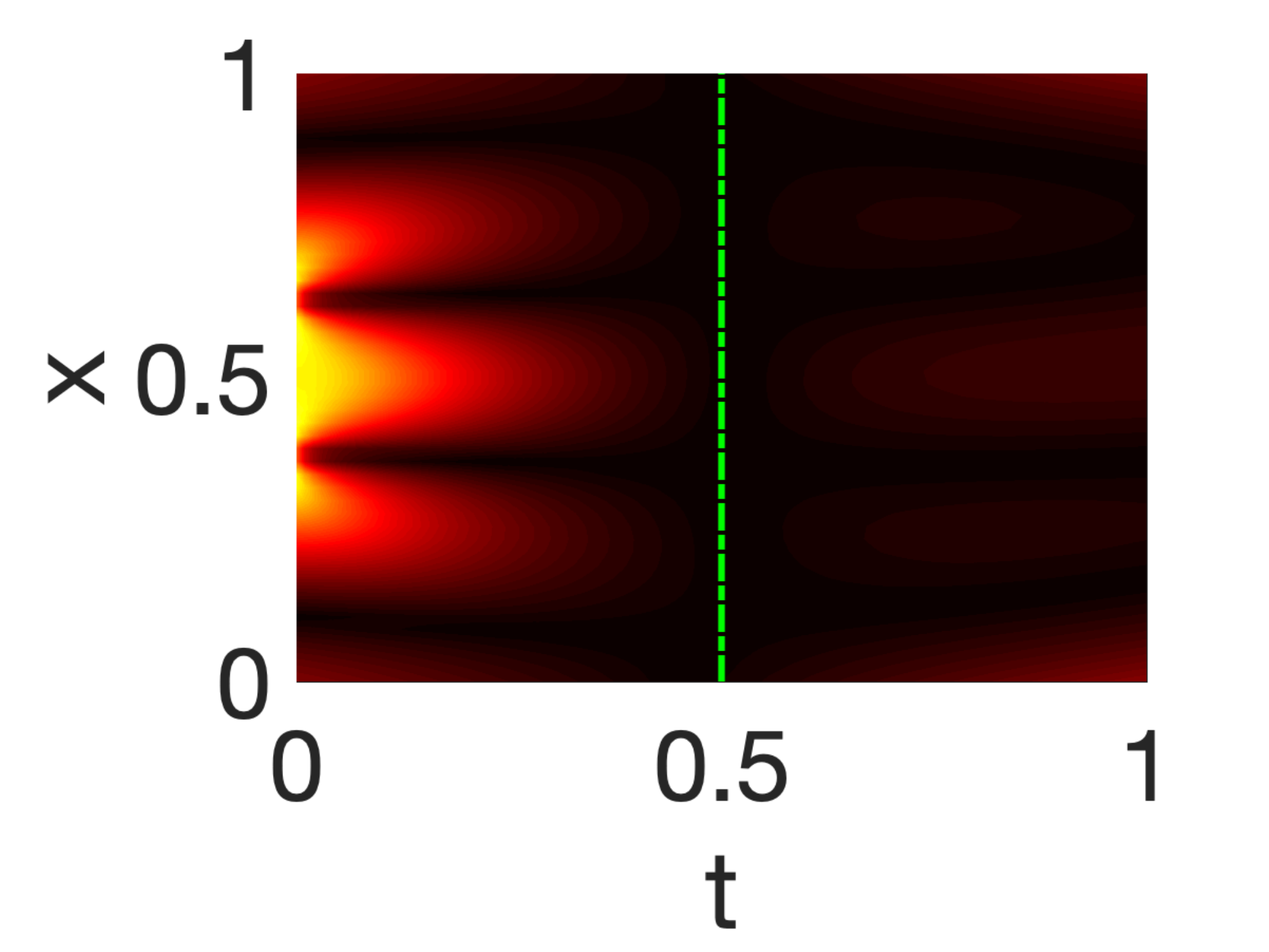}
			\caption{\small \ours} 
		\end{subfigure}
		\\
		\begin{subfigure}[b]{0.05\textwidth}
			\raisebox{0.4in}
			{
				\includegraphics[scale=0.17]{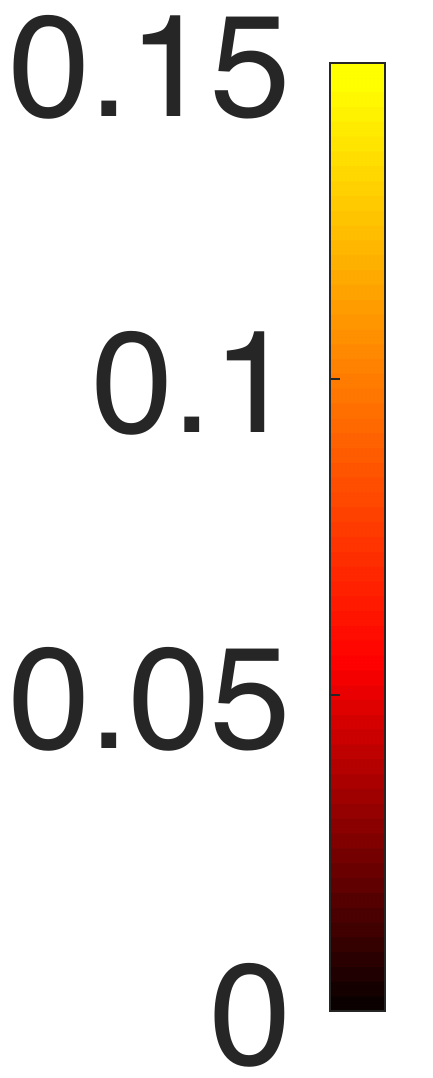}
			}
		\end{subfigure}
		\setcounter{subfigure}{4}
		&
		\begin{subfigure}[b]{0.24\textwidth}
			\centering
			\includegraphics[width=\linewidth]{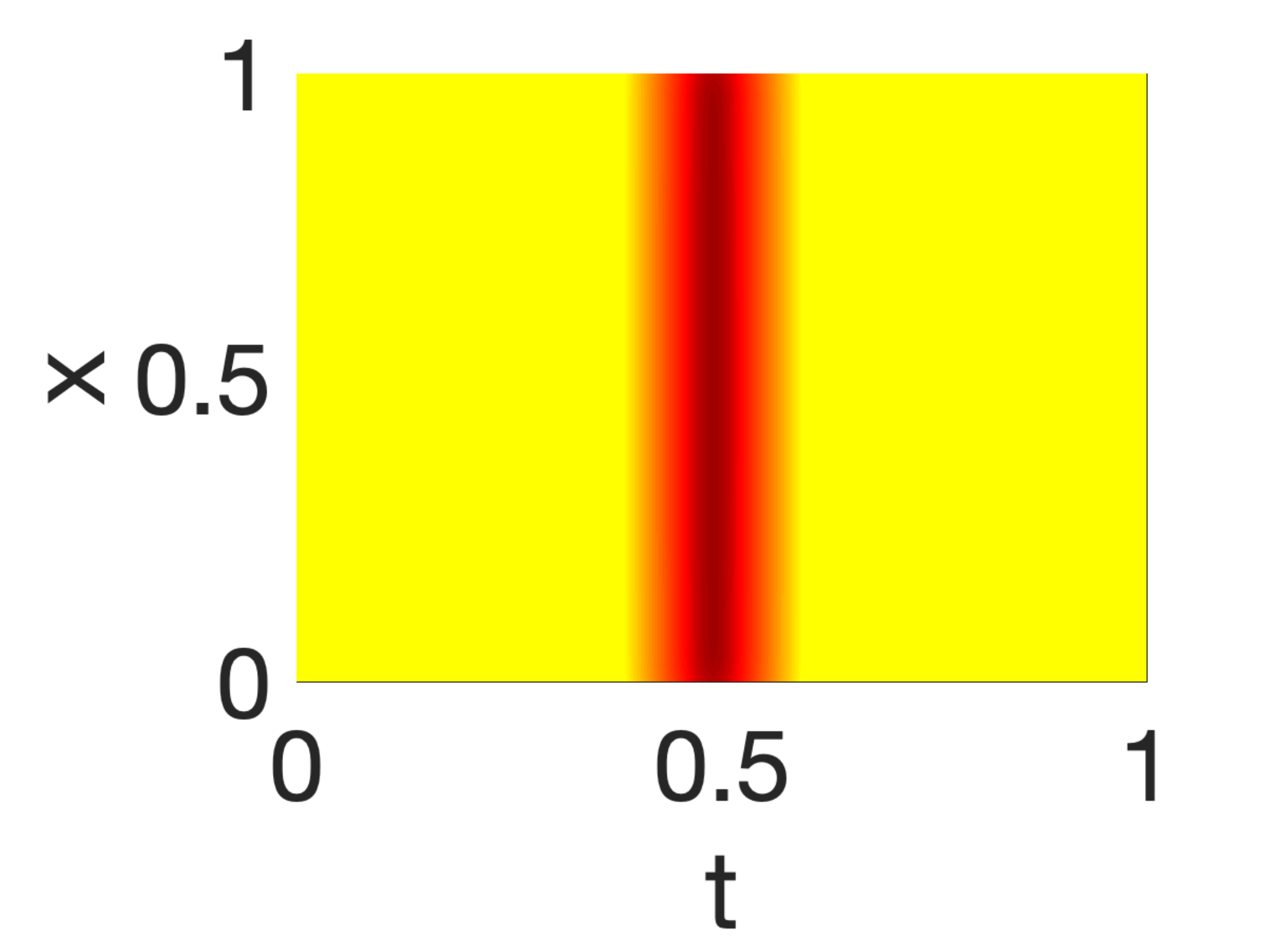}
			\caption{\small SKL (PSD)} 
		\end{subfigure}
		&
		\begin{subfigure}[b]{0.24\textwidth}
			\centering
			\includegraphics[width=\linewidth]{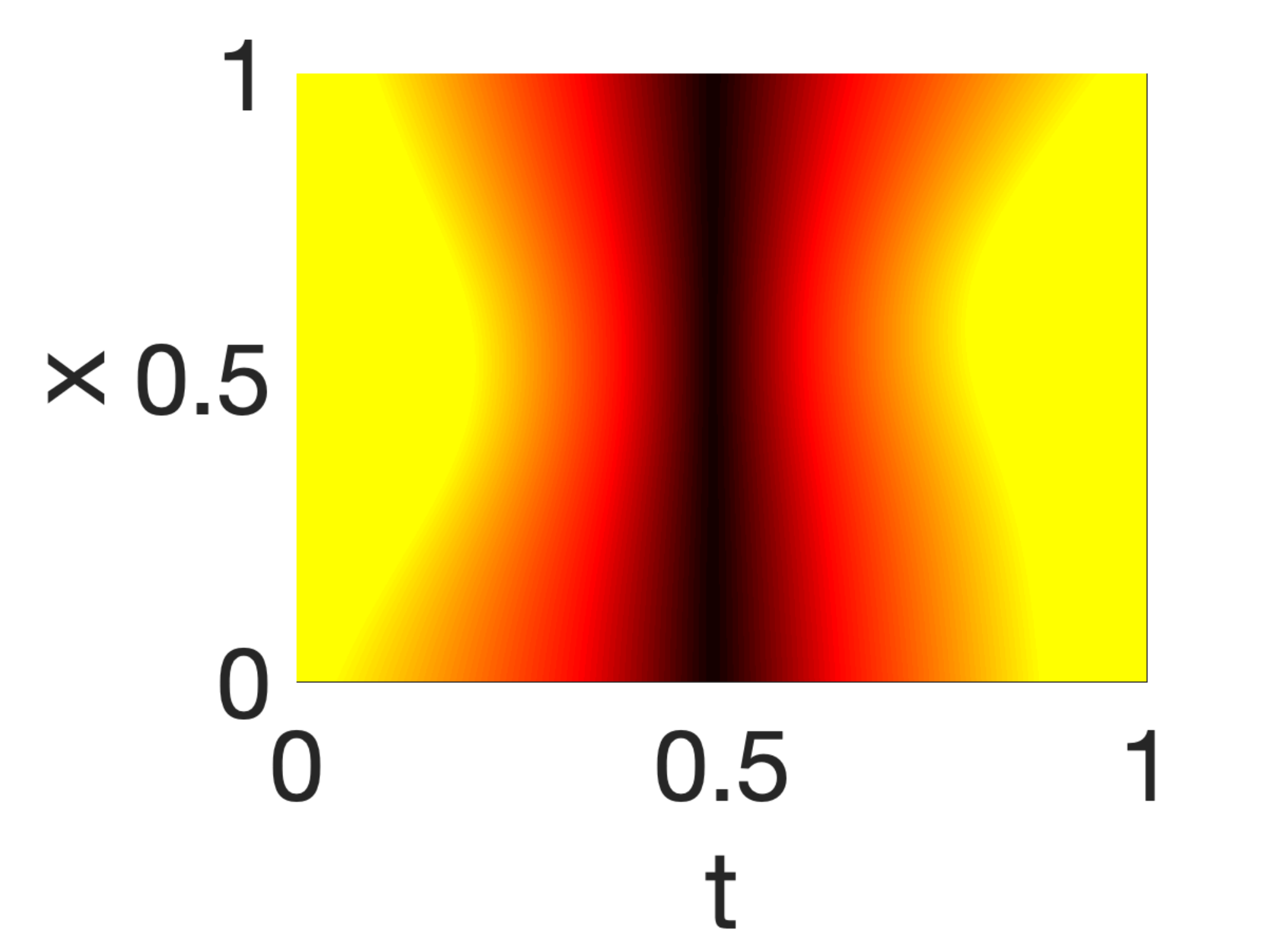}
			\caption{\small DKL (PSD)} 
		\end{subfigure}
		&
		\begin{subfigure}[b]{0.24\textwidth}
			\centering
			\includegraphics[width=\linewidth]{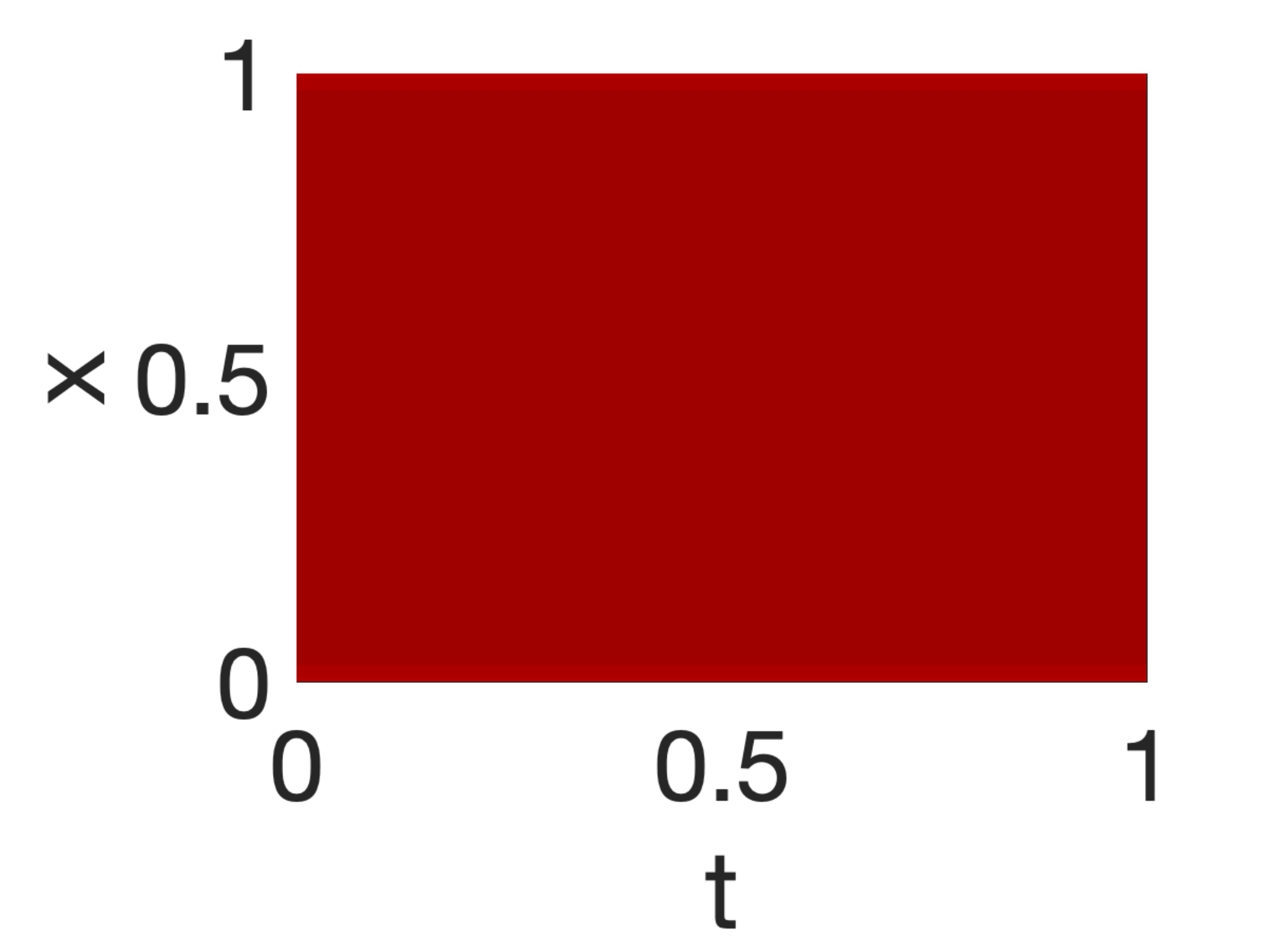}
			\caption{\small LFM (PSD)} 
		\end{subfigure}
		&
		\begin{subfigure}[b]{0.24\textwidth}
			\centering
			\includegraphics[width=\linewidth]{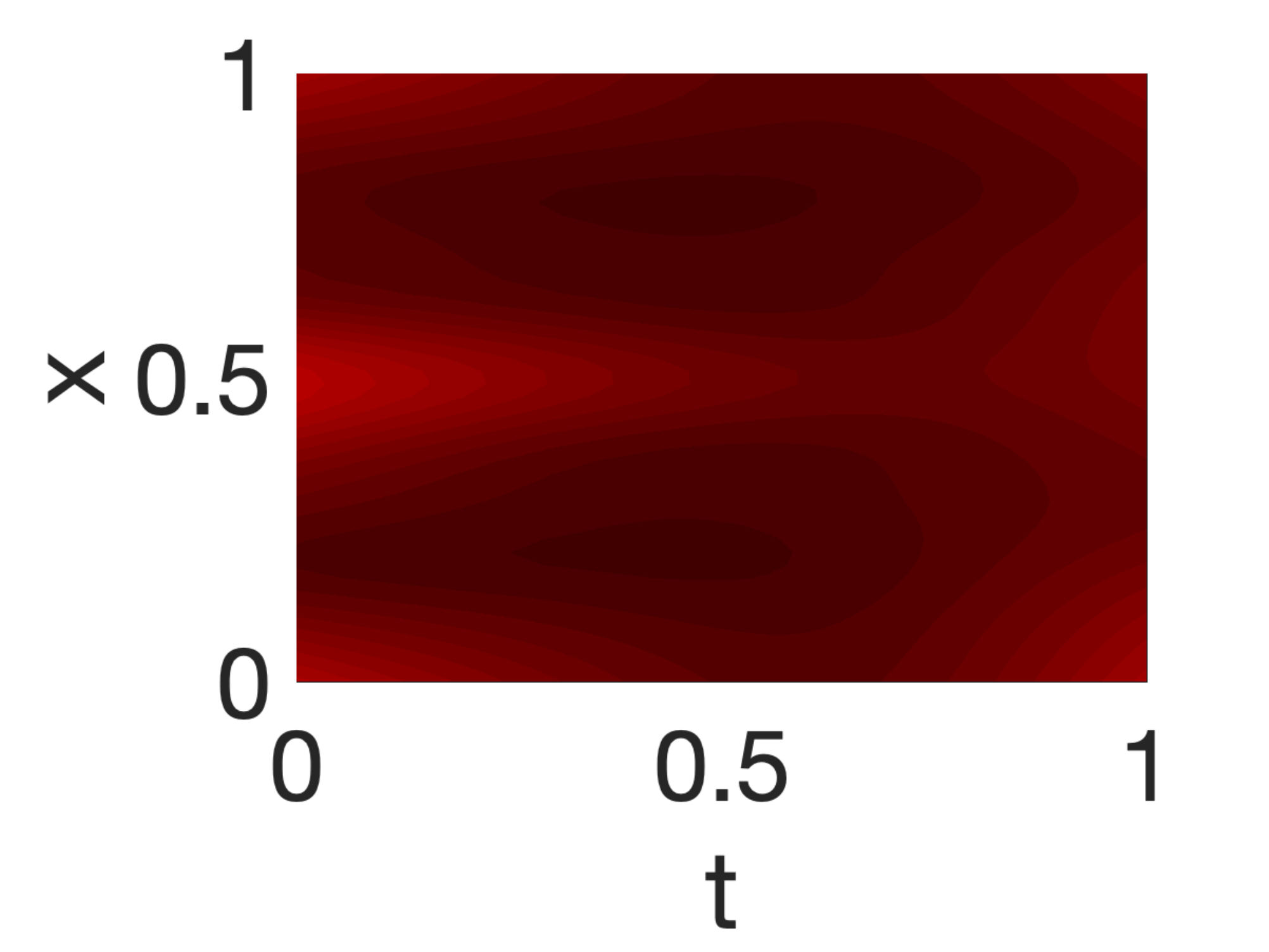}
			\caption{\small \ours (PSD)}
		\end{subfigure}
	\end{tabular}
	\vspace{-0.12in}
	%\caption{\small The absolute prediction error. The training examples stay on $t=0.5$ (the green line). }
	\caption{\small The prediction error (1st row) --- the absolute value of the difference between the posterior mean and the ground-truth  and  predictive uncertainty (2nd row) --- the posterior standard on \textit{1dDiffusion}. The training examples stay on $t=0.5$ (the green line). }
	\label{fig:1ddiffu}
	\vspace{-0.15in}
\end{figure*}

Since for diffusion equations, LFM cannot derive the kernel for time variable $t$ (the green function is for a single time variable only; see \citep{alvarez2009latent}),  for a fair comparison on \textit{1dDiffusion}, we fixed $t=0.5$ and used the $48$  spatial points as the training inputs. \cmt{While the kernel of LFM is for the spatial input only,  all the other methods used both the spatial and time inputs.}  We then evaluated the posterior distribution of the function values at all the grid points ($48 \times 101$) in the entire domain. We report the absolute difference between the posterior mean and ground-truth in Fig. \ref{fig:1ddiffu}a-d. We can see that the prediction errors of SKL/DKL are close to $0$ (dark colors) in regions close to the training data ($t=0.5$). However, when the test points are getting far away, say, close to the boundary ($t=0$ or $1$), the error grows significantly (see the bright colors). Overall, DKL still achieves smaller extrapolation error than SKL, implying an advantage of the more expressive deep kernel. From Fig. \ref{fig:1ddiffu}c, we can see that while LFM misses the time information, it still exhibits better extrapolation results, as compared with SKL/DKL, showing the benefit of the physics.  \ours achieves even smaller prediction error (\ie darker) when $t$ is away from the training time point and exhibits the best extrapolation performance. The RMSEs of all the methods are \{SKL: 0.18, DKL: 0.11, LFM: 0.09, \ours:0.07\}. We also report the predictive standard deviation  (PSD) of each method in Fig. \ref{fig:1ddiffu}e-f. We can see that the PSDs of SKL/DKL are both close to $0$ in the training region, and quickly increase when the inputs move away (on average DKL shows smaller PSDs and smoother changes). By  contrast, LFM and \ours obtain PSDs quite uniformly across the entire domain and less than SKL/DKL. It means that the physics knowledge help inhibit overfitting and reduce the uncertainty in extrapolation. Compared with LFM, \ours obtains even smaller PSDs (darker color) across the domain, showing even smaller uncertainty in extrapolation. %This is consistent with the results on \textit{1stODE}.  

\begin{figure*}
	\centering
	\setlength\tabcolsep{0pt}
	\begin{tabular}[c]{ccc}
		\begin{subfigure}[b]{0.33\textwidth}
			\centering
			\includegraphics[width=\linewidth]{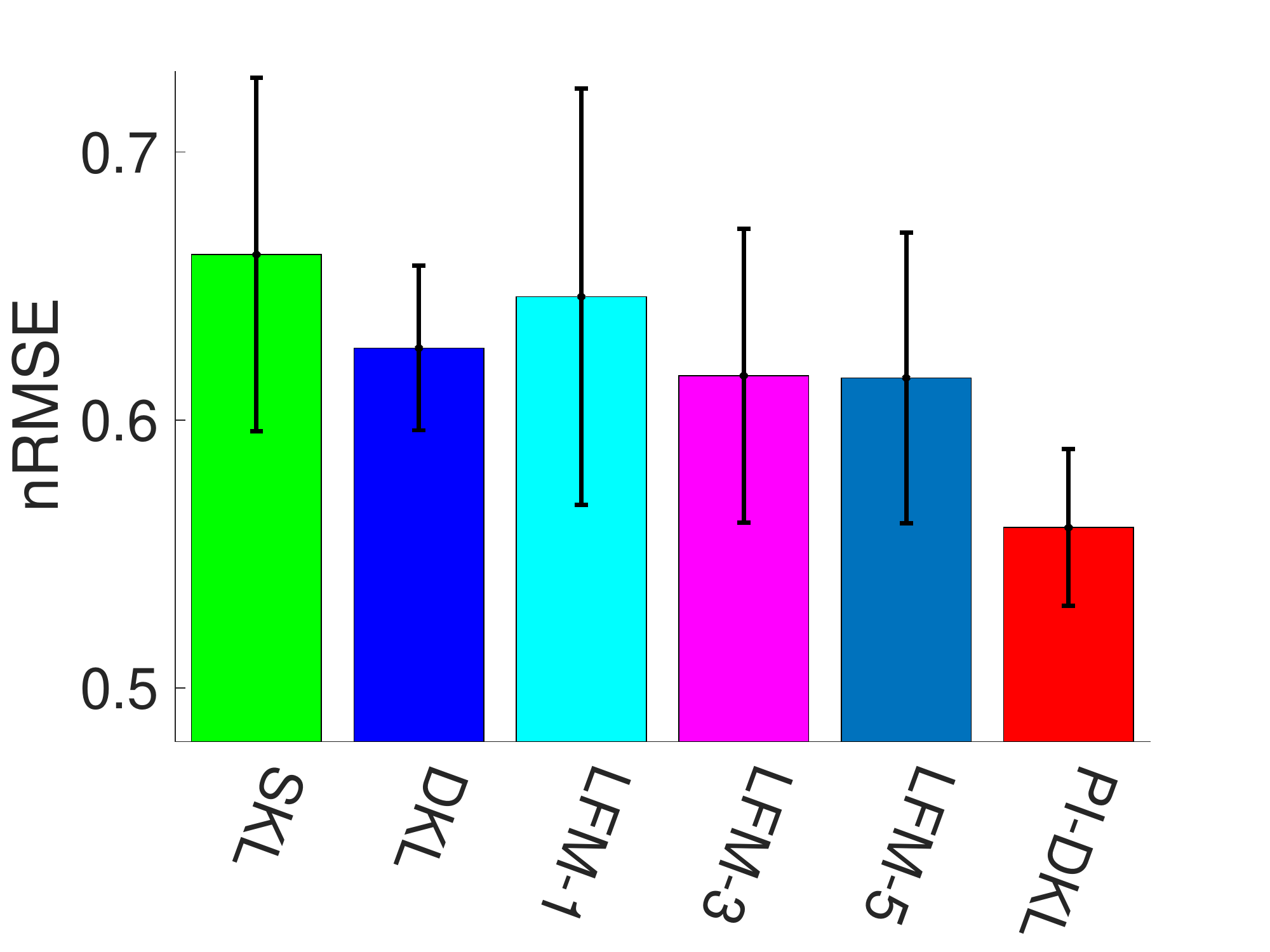}
			\caption{\small Copper} 
		\end{subfigure}
		&
		\begin{subfigure}[b]{0.33\textwidth}
			\centering
			\includegraphics[width=\linewidth]{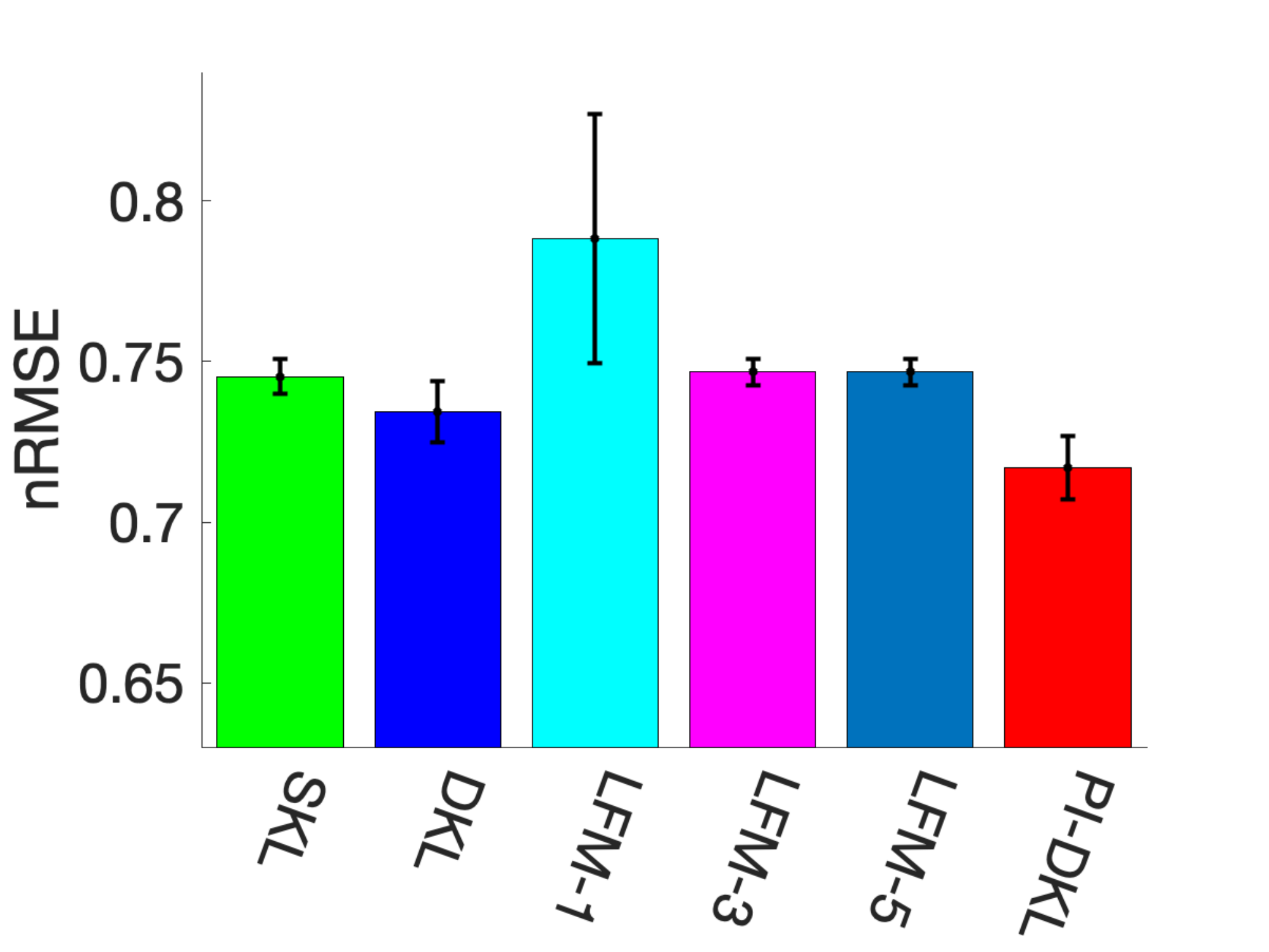}
			\caption{\small Cadmium} 
		\end{subfigure}
		&
		\begin{subfigure}[b]{0.33\textwidth}
			\centering
			\includegraphics[width=\linewidth]{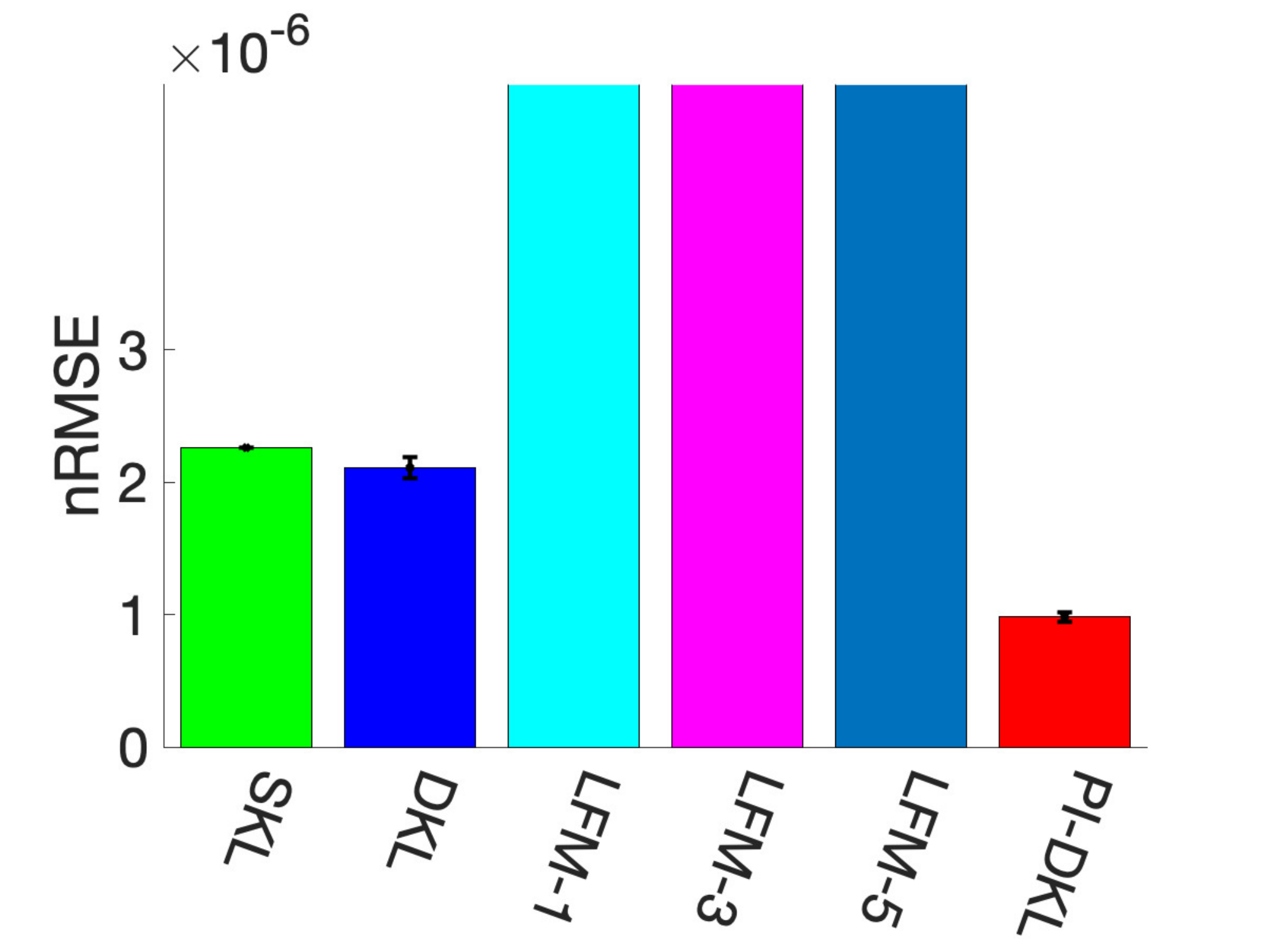} 
			\caption{\small Joint Motion} 
		\end{subfigure}
	\end{tabular}
	\vspace{-0.15in}
	\caption{\small Metal concentration prediction in Swiss Jura (a, b) and joint angle prediction in motion capture (c). The results are averaged over $5$ runs. The normalized root-mean-square error (nRMSE) in each run is computed by normalizing the RMSE by the mean of the test outputs. }
	\vspace{-0.25in}
	\label{fig:jura-motion}
\end{figure*}
%\vspace{-0.05in}
\subsection{Real-World Applications}
\vspace{-0.05in}
%\textbf{Metal Pollution in Swiss Jura.} 
\subsubsection{Metal Pollution in Swiss Jura} 
\vspace{-0.05in}
Next, we evaluated \ours in real-world applications. We examined the predictive performance in terms of normalized RMSE (nRMSE) and test log-likelihood (LL). Due to the space limit, the test LL results are provided in the supplementary material. 
We first considered predicting the metal concentration in Swiss Jura. The data were collected from  300 locations in a 14.5 km$^2$ region~(\url{https://rdrr.io/cran/gstat/man/jura.html}). 
The diffusion of the metal concentration is naturally modeled by a diffusion equation with the two-dimensional spatial domain, $\frac{\partial f(x_1, x_2, t)}{\partial t} = \alpha (\frac{\partial f^2(x_1,x_2, t)}{\partial x_1^2} +\frac{\partial f(x_1, x_2, t)}{\partial x_2^2})$,
where $f(\cdot, \cdot, \cdot)$ is the concentration of the metal at a particular location and time point.  However, the dataset do not include the time $t_s$ when these concentrations were measured. LFM considers the initial condition $f(x_1, x_2,0)$ as the latent source and obtains a kernel of the locations where $t_s$ can be viewed as a kernel parameter learned from data. In our approach, we estimated the solution function at $t_s$, $h(x_1, x_2) =f(x_1, x_2, t_s)$. Hence, the equation can be viewed as 
\begin{align}
\frac{\partial h^2(x_1,x_2)}{\partial x_1^2} +\frac{\partial h^2(x_1, x_2)}{\partial x_2^2} = g(x_1, x_2), \notag
\end{align}
where the latent source $g(x_1, x_2) = \frac{1}{\alpha}\frac{\partial f(x_1, x_2, t)}{\partial t}|_{t = t_s} $. 
We were interested in predicting the concentration of cadium and copper. The input variables include the coordinates of the location $(x_1, x_2)$, the concentrations of \{nickel, zinc\} for cadmium, and \{lead, nickel, zinc\} for copper. For \ours, we selected $m$ from $\{10, 50, 100, 200, 500\}$ for the generative component and $\gamma$ from $\{0.01, 0.05, 0.1, 0.5, 1, 2, 5, 10\}$. We normalized the training inputs and then sampled $\Z$ from $\N(\0,\I)$ in model estimation. 
For LFM, we varied the number  of latent forces from \{1,3, 5\}. 
We randomly selected 50 samples for training, and used the remaining 250 samples for test. We repeated the experiments for 5 times, and report the average nRMSE and its standard deviation of each method in Fig. \ref{fig:jura-motion}a and b. 
 \ours outperforms all the competing approaches for both prediction tasks. \ours always significantly improves upon SKL and DKL ($p<0.05$). In addition, \ours significantly outperforms LFM in predicting Cadium concentration (Fig. \ref{fig:jura-motion}b). Note that LFM does improve upon SKL in predicting Copper concentration (Fig. \ref{fig:jura-motion}a), but not as significant as \ours. % advantage over GP-ARD, even with 5 latent forces. Its performance might be further improved by jointly learning to predict the concentrations of multiple metals (\ie multi-output regression). 

%\textbf{Motion Capture.} 
\subsubsection{Motion Capture}
\vspace{-0.05in}
We then looked into predicting  trajectories of joints in the motion capture application. To this end, we used CMU motion capture database ( \url{http://mocap.cs.cmu.edu/}), from which we used the samples collected from subject 35 in the walk and jog motion lasting for 2,644 seconds. We trained all the models to predict the angles of Joint 60 along with time.  We used the first order ODE  in \eqref{eq:ode} to represent the physical model, based on which we ran LFM and \ours.  Note this physical system might be oversimplified~\citep{alvarez2009latent}. For LFM, we varied the number of latent forces from \{1,3, 5\}. Again, we randomly selected $500$ samples for training and $2,000$ samples for test. We repeated the experiments for $5$ times and report the average nRMSE and its standard deviation in Fig. \ref{fig:jura-motion}c.  As we can see, \ours improves upon all the competing methods by a large margin. Note that LFM is even far worse than SKL. This might because LFM over-exploits the over-simplified physics, which harms the prediction. By contrast, \ours allows us to tune the number of virtual observations $m$ and the likelihood weight ($\gamma$ in \eqref{eq: elbo}), and hence can consistently improve upon DKL. 

%\begin{figure}%{r}{0.5\textwidth}	
\vspace{-0.05in}
\subsubsection{PM2.5 in Salt Lake City} \label{exp:pm25}
%\vspace{-0.08in}
\begin{figure}[H] %{r}{0.5\textwidth}
\vspace{-0.2in}
	\setlength\tabcolsep{0.0pt}
	\begin{tabular}[c]{cc}
		\begin{subfigure}[b]{0.25\textwidth}
			\centering
			\includegraphics[width=\textwidth]{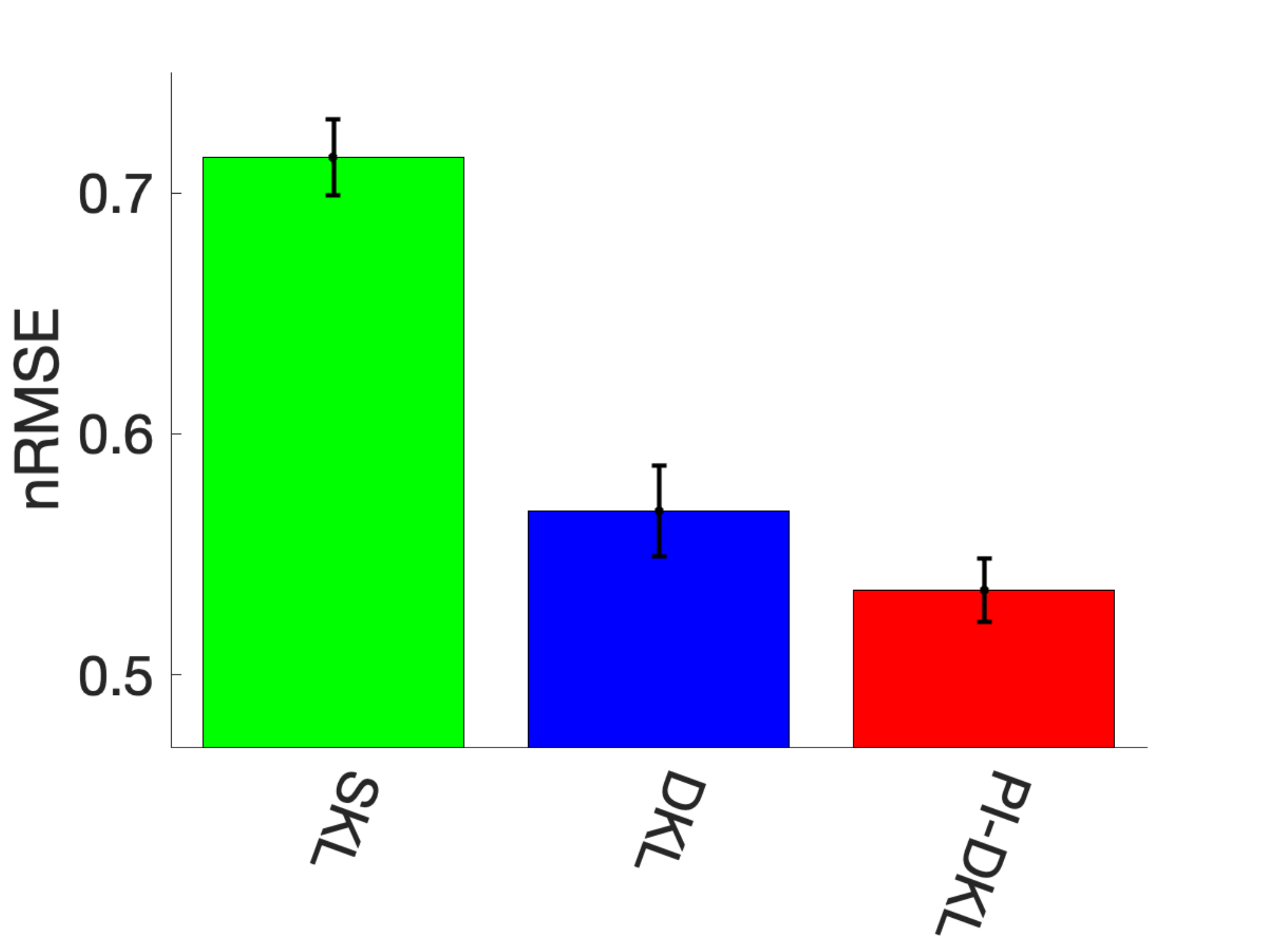}
			\vspace{-0.05in}
			\caption{\small PM2.5}
		\end{subfigure} & 
		\begin{subfigure}[b]{0.25\textwidth}
			\centering
			\includegraphics[width=\textwidth]{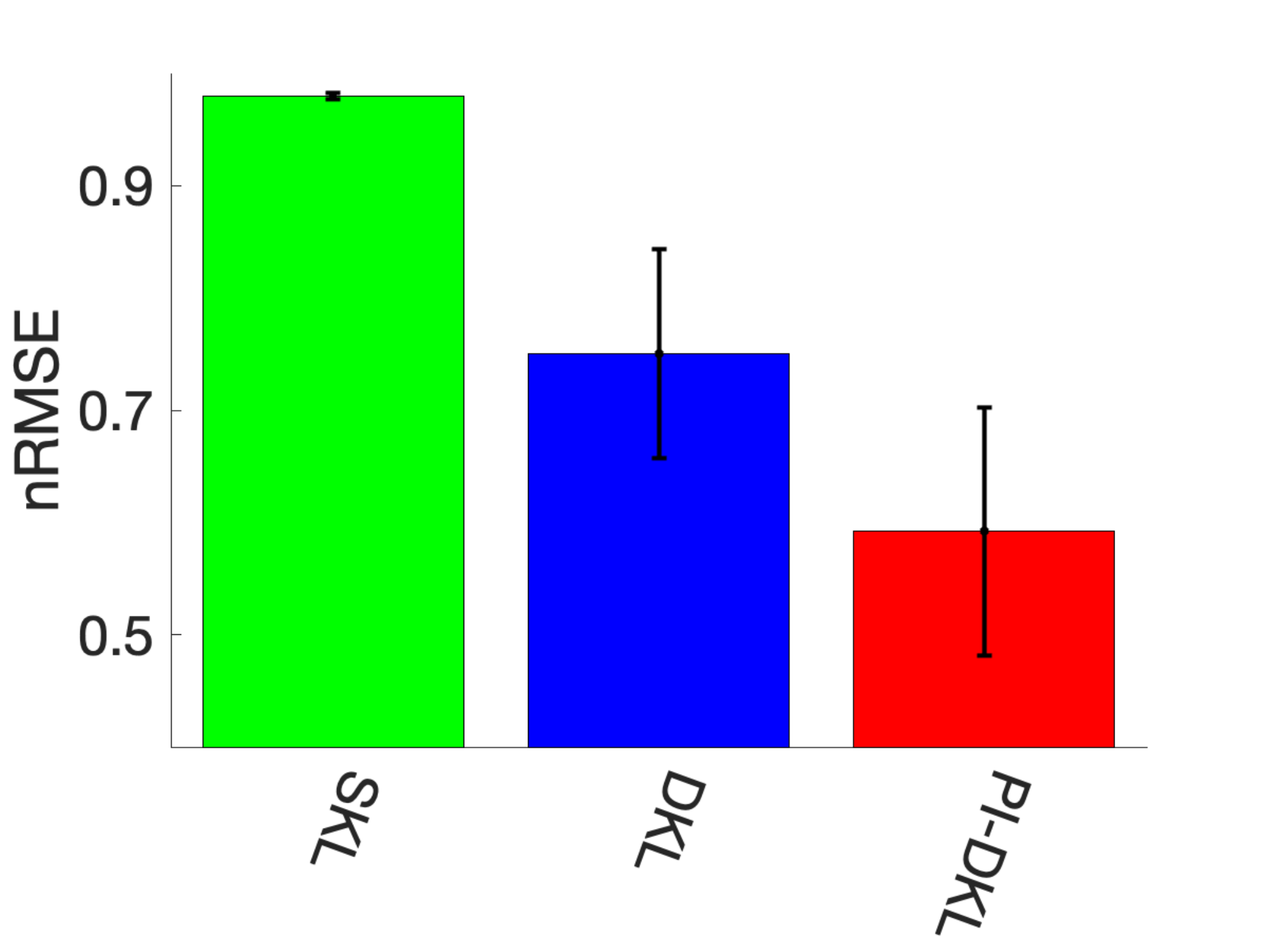}
			\vspace{-0.05in}
			\caption{\small Traffic flow}
		\end{subfigure}
	\end{tabular}
	\vspace{-0.15in}
	\caption{\small PM2.5 and traffic flow prediction.} \label{fig:pm-traffic}
\vspace{-0.2in}
\end{figure}

%\textbf{PM2.5 in Salt Lake City.} 
Next, we considered predicting the Particulate Matter (PM2.5) levels across Salt Lake City. The dataset were collected from sensors' reads at different time and locations (\url{https://aqandu.org/}). We chose the time range from 07/04/2018 to 07/06/2018.  Following~\citep{wang2018prediction}, we used the diffusion equation plus a latent source term to represent the physical model,
\[
\frac{\partial f(x_1, x_2, t)}{\partial t} - \alpha \sum_{j=1}^2\frac{\partial f^2(x_1,x_2, t)}{\partial x_j^2}  = g(x_1, x_2, t),
\]
where $f$ is the concentration level and $g$ the unknown source. 
The input variables  include both the location coordinates and detailed time points. Since LFM cannot construct a full kernel of the input variables from the physics, we did not test it to avoid unfair comparisons. We trained SKL and DKL with both the spatial and time inputs. We randomly selected $500$ samples for training and $2,000$ samples for test. We repeated the experiments for $5$ times and report the average nRMSE and its standard deviation in Fig. \ref{fig:pm-traffic}a. As we can see, with a more expressive kernel, DKL improves upon SKL significantly, and with the incorporation of the physics, \ours in turn outperforms DKL significantly ($p< 0.05$).  

\vspace{-0.05in}
\subsubsection{High-Way Traffic Flow Prediction}\label{exp:traffic}
\vspace{-0.05in}
Finally, we applied \ours to predict the traffic flow in the interstate highway 215 across Utah state.  The Utah Department of Transportation (UDOT) has installed sensors every a few miles along the high way. Each sensor counts the number of vehicles passed every minute, and sends the data back to a central database. The real time data and road conditions are available at \url{https://udot.iteris-pems.com/}. We used the data collected by 20 sensors continuously installed in a segment of $30$ miles, and the time was chosen from 08/05/2019 to 08/11/2019.  The input variables include the location coordinates of each sensor and the time of each read. Following~\citep{nagatani2000density}, we used the Burgers' equation plus a latent source term to describe the system,  
\[
\frac{\partial f}{\partial t} + f\cdot \sum_{j=1}^2\frac{\partial f}{\partial x_i}- \nu \sum_{j=1}^2\frac{\partial f^2}{\partial x_j^2}  = g(x_1, x_2, t),
\]
where $f$ is the traffic flow, $\nu$ the unknown viscous coefficient,  and $g$ the latent source. The equation is nonlinear and we do not have an analytical form of Green's function; the source $g$ is a function of both time and spatial inputs. Hence we cannot use LFM to incorporate the physics to enhance GP training, and we only compared with SKL and DKL. We randomly selected $500$ and $2,000$ samples for training and test, respectively, and repeated for $5$ times. The average nRMSEs and the standard deviations are reported in Fig. \ref{fig:pm-traffic}b. As we can see, DKL significantly outperforms SKL, which demonstrates the advantage of the more expressive deep kernel. More important, \ours further improves upon  DKL, showing that the physics incorporated by our approach indeed promotes the prediction accuracy.
In all these real datasets, the predictive performance in terms of test LL that integrates uncertainty shows consistent results (see the supplementary material).

\vspace{-0.05in}
\section{Conclusion}
\vspace{-0.05in}
%\vspace{-0.15in}
We proposed \ours, a physics informed deep kernel learning that can flexibly incorporate physics knowledge from incomplete differential equations to improve function learning and uncertainty quantification. In the future, we will extend our model with sparse approximations~\citep{GPSVI13,wilson2016stochastic} to exploit  physics in large-scale applications and for multi-output regression tasks.

\bibliographystyle{apalike}
\bibliography{PIGP}

\section*{Supplementary Materials}%\maketitle

\section*{Test Log-likelihood on Real-World Datasets}
\begin{figure*}[htb!]
	\centering
	\setlength\tabcolsep{0pt}
	\begin{tabular}[c]{ccc}
		\begin{subfigure}[b]{0.33\textwidth}
			\centering
			\includegraphics[width=\linewidth]{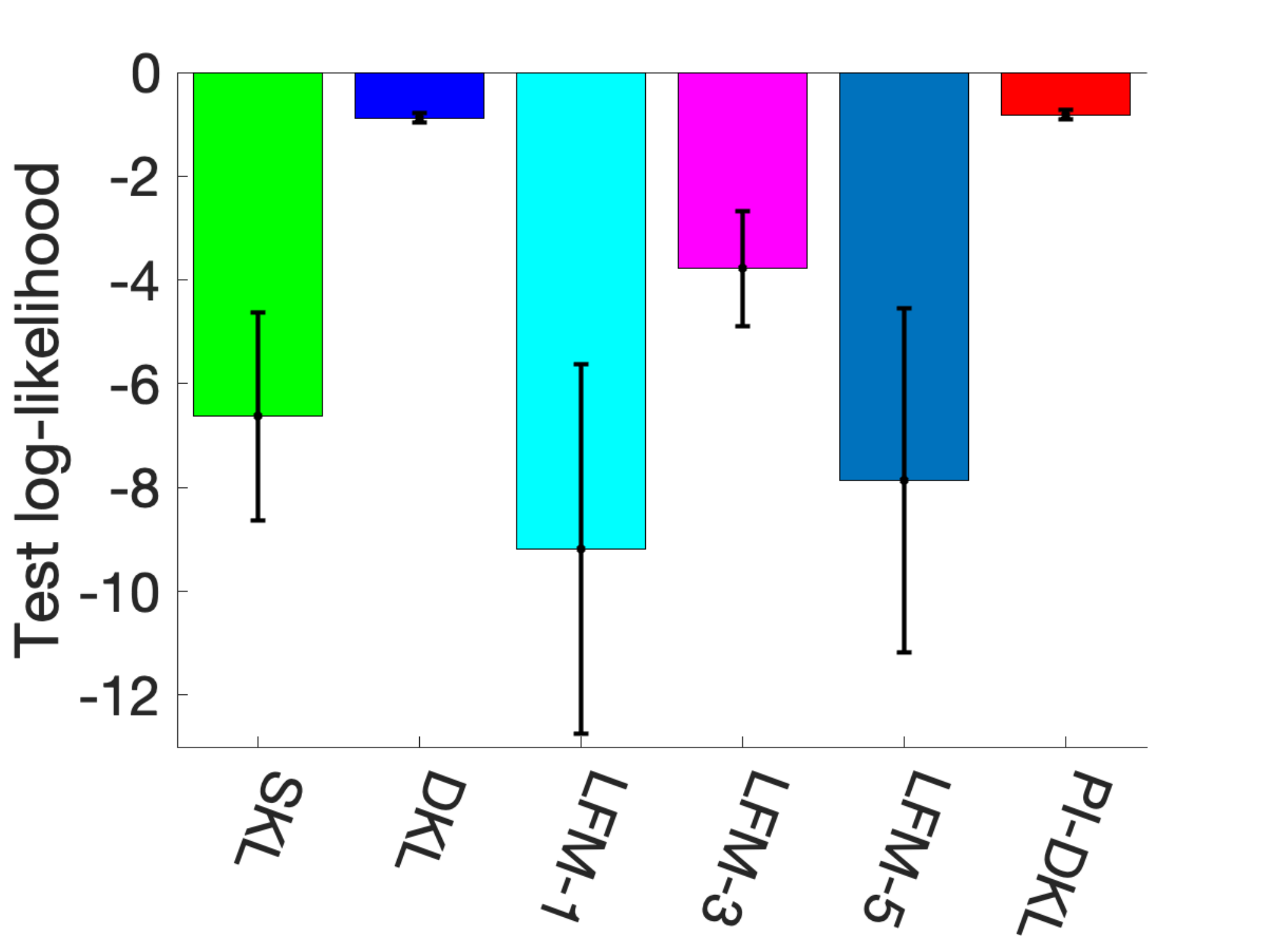}
			\caption{\small Copper} 
		\end{subfigure}
		&
		\begin{subfigure}[b]{0.33\textwidth}
			\centering
			\includegraphics[width=\linewidth]{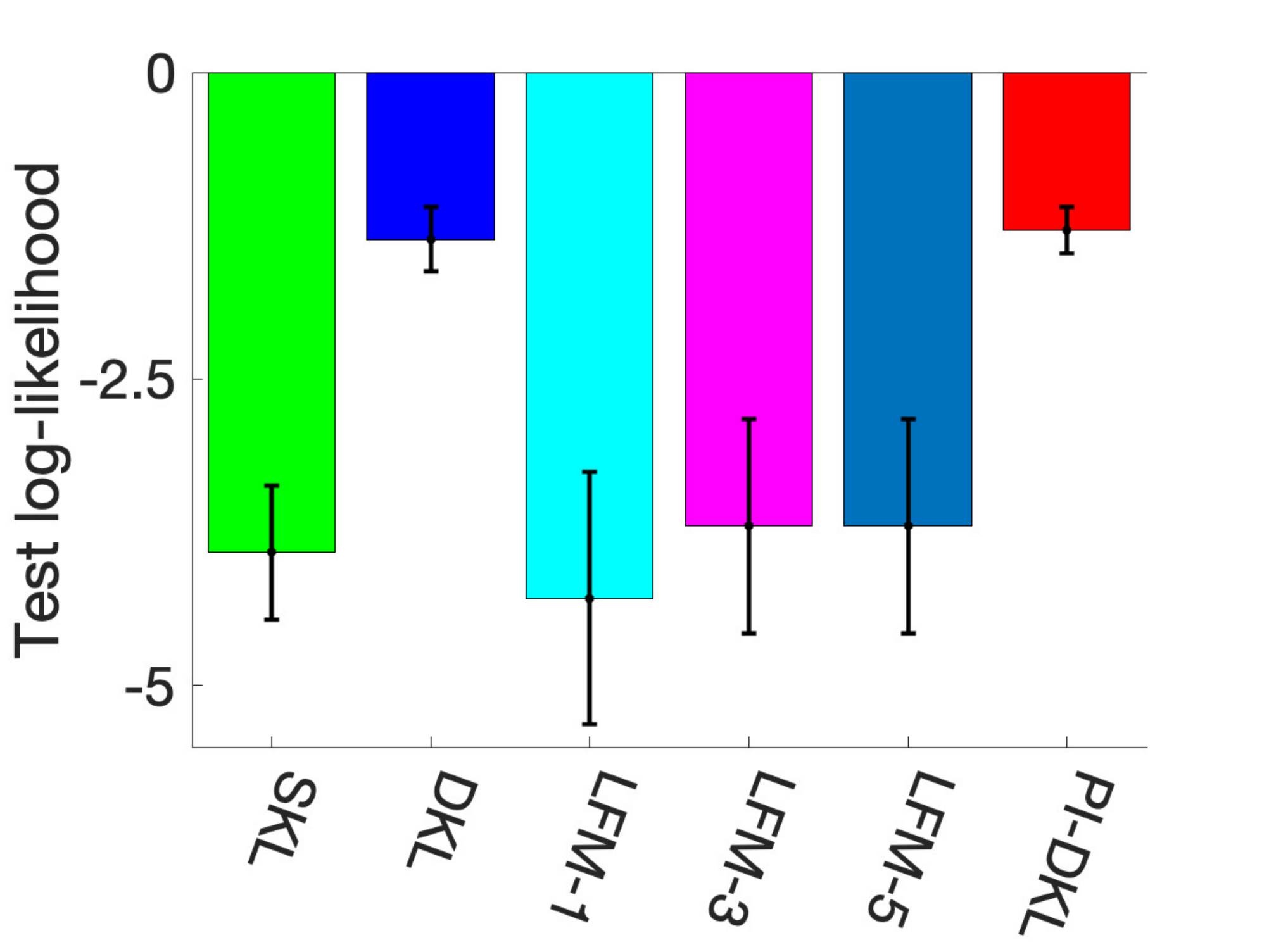}
			\caption{\small Cadmium} 
		\end{subfigure}
		&
		\begin{subfigure}[b]{0.33\textwidth}
			\centering
			\includegraphics[width=\linewidth]{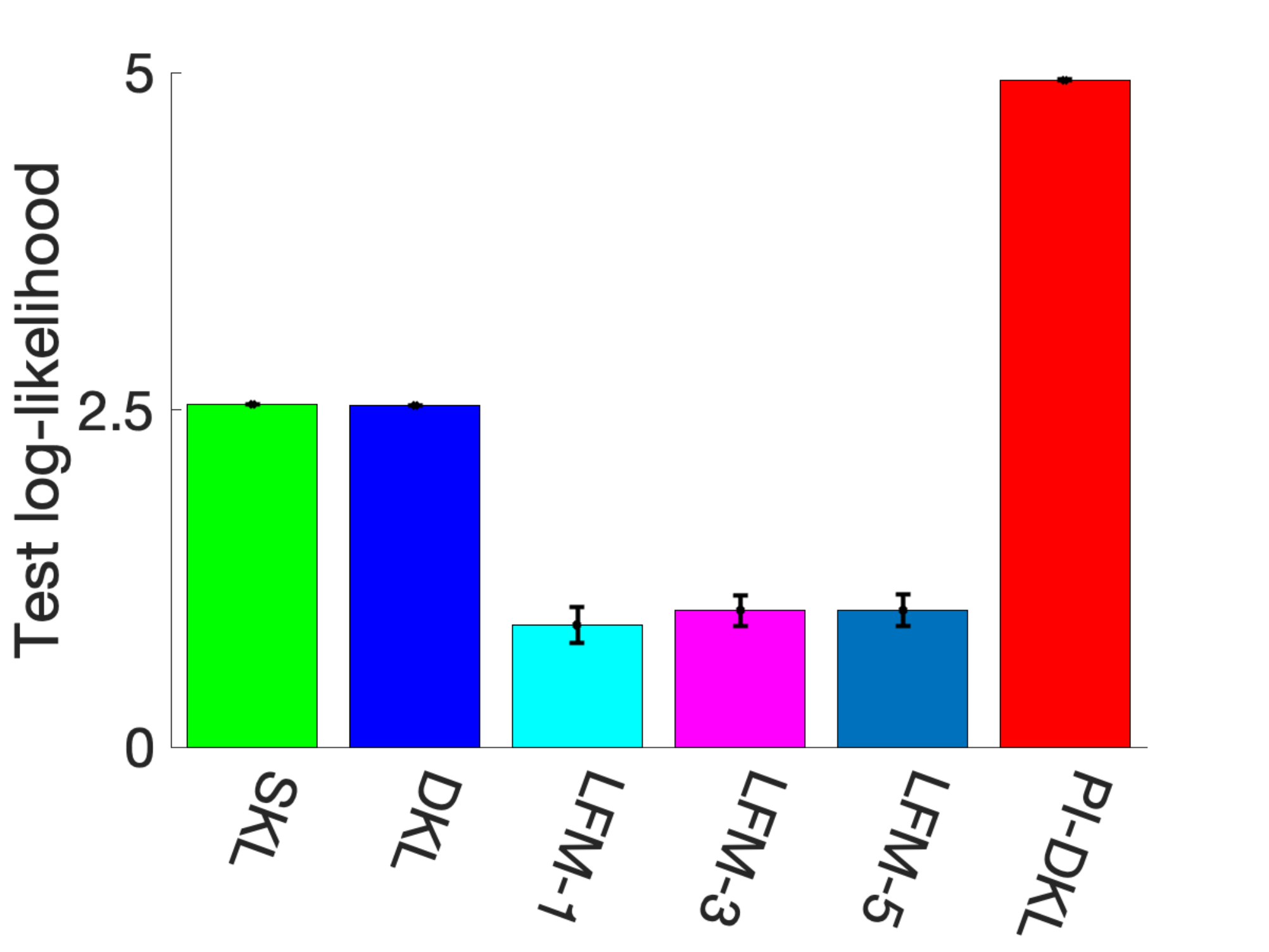} 
			\caption{\small Motion} 
		\end{subfigure}
	\end{tabular}
	\vspace{-0.15in}
	\caption{\small Test log-likelihood (LL) in Swiss Jura (a, b) and joint angle prediction in motion capture (c). The results are averaged over $5$ runs. }
	\vspace{-0.25in}
	\label{fig:jura-motion-ll}
\end{figure*}

\begin{figure*}[htb!]
	\centering
	\setlength\tabcolsep{0pt}
	\begin{tabular}[c]{ccc}
		\begin{subfigure}[b]{0.33\textwidth}
			\centering
			\includegraphics[width=\linewidth]{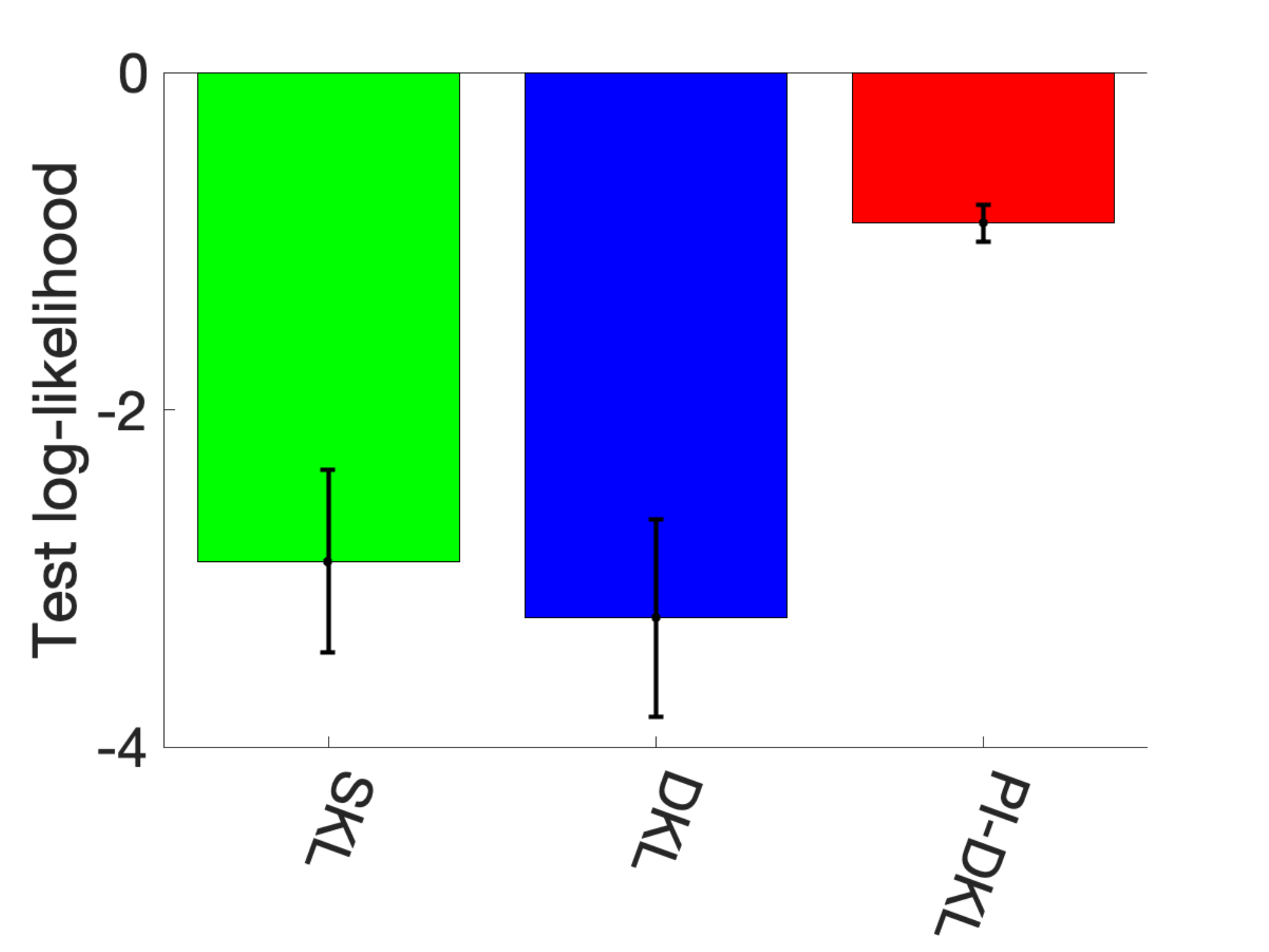}
			\caption{\small Copper} 
		\end{subfigure}
		&
		\begin{subfigure}[b]{0.33\textwidth}
			\centering
			\includegraphics[width=\linewidth]{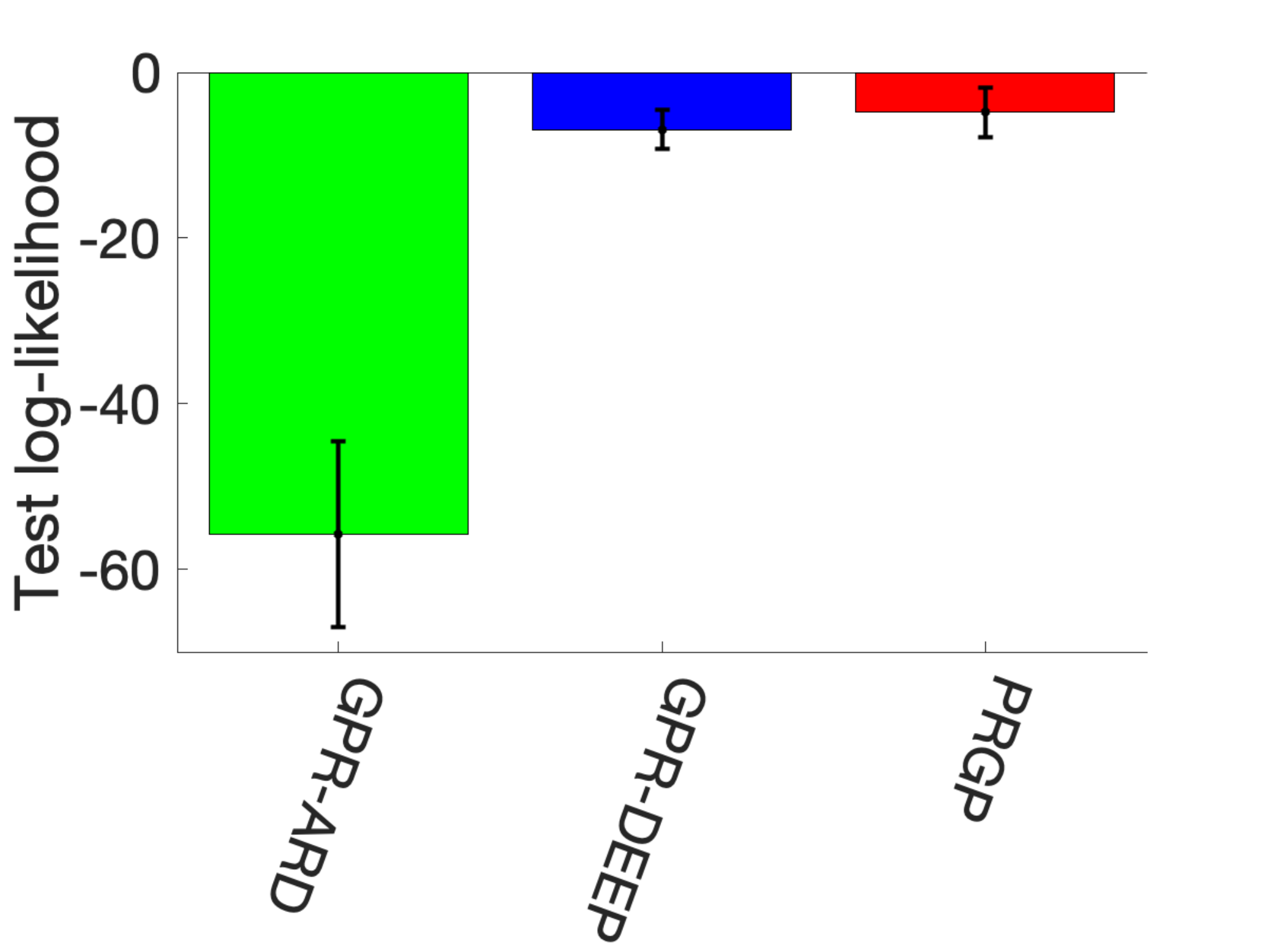}
			\caption{\small Cadmium} 
		\end{subfigure}
		&

	\end{tabular}
	\vspace{-0.15in}
	\caption{\small Test log-likelihood (LL) for PM2.5 and traffic flow datasets. The results are averaged over $5$ runs. }
	%\vspace{-0.25in}
	\label{fig:pm25-ll}
\end{figure*}

In Fig. \ref{fig:jura-motion-ll} and \ref{fig:pm25-ll}, we report the test log-likelihood (LL) of all the methods in the real-world applications in Section 6.2 of the main paper. Note that since test LLs are negative (smaller than zero) in most datasets, the corresponding bar plots are shown inverted for a convenient comparison. As we can see, our method (PI-DKL) consistently outperforms all the competing methods, and in many cases by a large margin. DKL  always obtains test LLs larger than or comparable  to SKL except that in Fig. \ref{fig:pm25-ll} a, DKL is slightly worse. It demonstrates the advantage of more expressive kernels. PI-DKL further improves upon DKL in all the cases, showing that the physics knowledge are effectively exploited and indeed improves prediction. Especially, in Fig. \ref{fig:pm25-ll}a, while DKL obtains slightly smaller test LLs than SKL, after PI-DKL regularizes the same deep kernel with physics, the test LLs are greatly improved. Note that, similar to nRMSE results, we can see LFM improves upon SKL in some cases, \eg LFM-3 in Fig. \ref{fig:jura-motion-ll} a and b, but in other cases are even worse, \eg in Fig. \ref{fig:jura-motion-ll} c. This might because the rigid incorporation (hard-coding) of the physics in LFM can even hurt the performance when there is a significant mismatch to the actual data. For example, a first-order ODE might be too simple to describe the motion data in Fig. \ref{fig:jura-motion-ll} c. Overall, the test LL results are consistent with nRMSEs shown in the main paper. 

\section*{Marginal Distribution of $\g$}
We have $\h = [h(\z_1, \epsilon), \ldots, h(\z_m, \epsilon)]$, where $h(\cdot, \epsilon) = \psi[\mu(\cdot) + \epsilon \sqrt{v(\cdot)}]$ (see (6) of the main paper). To  obtain each element $\tg_j$ in $\g$, we can first sample the Gaussian random noise, $\epsilon \sim \N(\epsilon|0, 1)$, and sample $\tg_j  \sim \delta\left(\tg_j - h(\z_j, \epsilon)\right) $ (see (7) of the main paper). However, we can also consider the marginal distribution each $\tg_j$. Since  $\tg_j$ is a transformation of Gaussian noise $\epsilon$, $\tg_j = \alpha_j(\epsilon)$ where $\alpha_j(\cdot) = h(\z_j, \cdot)$. The marginal distribution of $\tg_j$ is 
\begin{align}
	p(\tg_j|\X, \y) = \N\left(\alpha_j^{-1}(\tg_j)|0, 1\right)|\frac{\d \epsilon}{\d \tg_j}|.
\end{align}
Although conceptually available, the marginal distribution is tricky to compute ---  the transformation $\alpha_j(\cdot)$ couples complex differential operators in $\psi$ and nonlinear functions $\mu(\cdot)$ and $v(\cdot)$. The inverse $\alpha_j^{-1}(\cdot)$ is very complicated and likely to have no closed-forms. The marginal joint distribution for $\g$ will be even more difficult to compute. Therefore, we choose to explicitly sample $\epsilon$ and then obtain the sample for $\g$ accordingly from the Delta prior in (7) of the main paper.

\end{document}